\documentclass{article}

\usepackage{PRIMEarxiv}

\usepackage[utf8]{inputenc} 
\usepackage[T1]{fontenc}    
\usepackage{hyperref}       
\usepackage{url}            
\usepackage{booktabs}       
\usepackage{amsfonts}       
\usepackage{nicefrac}       
\usepackage{microtype}      
\usepackage{lipsum}
\usepackage{fancyhdr}       
\usepackage{graphicx}       
\usepackage{amsmath}
\graphicspath{{media/}}     

\usepackage{algorithm}
\usepackage{algpseudocode}
\usepackage{multirow}
\title{AgriPINN: A Process-Informed Neural Network for Interpretable and Scalable Crop Biomass Prediction Under Water Stress
\thanks{\textit{\underline{Citation}}: 
\textbf{Authors. Title. Pages.... DOI:000000/11111.}} 
}

\author{Yue Shi,  Liangxiu Han*, Xin Zhang, Tam Sobeih\\
  Department of Computing, and Mathematics,\\
  Faculty of Science and Engineering,  \\
  Manchester Metropolitan University, \\
  Manchester, M1 5GD, UK. \\
  \texttt{\{y.shi, l.han\}@mmu.ac.uk} \\
   \And
  Amit Kumar Srivastava, Krishnagopal Halder, Frank Ewert \\
  Leibniz Centre for Agricultural Landscape Research (ZALF)\\
  Eberswalder Str \\
  Müncheberg, Germany\\
  \texttt{email@email} \\
   \And
  Thomas Gaiser, Nguyen Huu Thuy, Dominik Behrend \\
  Institute of Crop Science and Resource Conservation (INRES)\\
  University of Bonn,\\
  Bonn, Germany.\\
  \texttt{tgaiser@uni-bonn.de} \\
}

\begin{document}
\maketitle

\begin{abstract}
Accurate prediction of crop above-ground biomass (AGB) under water stress is critical for monitoring crop productivity, guiding irrigation, and supporting climate-resilient agriculture. Data-driven models scale well but often lack interpretability and degrade under distribution shift, whereas process-based crop models (e.g. DSSAT, APSIM, LINTUL5) require extensive calibration and are difficult to deploy over large spatial domains. To address these limitations, we propose AgriPINN, a process-informed neural network that integrates a biophysical crop-growth differential equation as a differentiable constraint within a deep learning backbone. 
This design encourages physiologically consistent biomass dynamics under water-stress conditions while preserving model scalability for spatially distributed AGB prediction. AgriPINN recovers latent physiological variables, including leaf area index (LAI), absorbed photosynthetically active radiation (PAR), radiation use efficiency (RUE), and water-stress factors, without requiring direct supervision. We pretrain AgriPINN on 60 years of historical data across 397 regions in Germany and fine-tune it on three years of field experiments under controlled water treatments. Results show that AgriPINN consistently outperforms state-of-the-art deep-learning baselines (ConvLSTM-ViT, SLTF, CNN-Transformer) and the process-based LINTUL5 model in terms of accuracy (RMSE reductions up to $43\%$) and computational efficiency (8× faster). Moreover, the latent variables estimated by AgriPINN align with independent observations, enabling interpretable assessment of water-stress dynamics and regional variation in AGB. By combining the scalability of deep learning with the biophysical rigor of process-based modeling, AgriPINN provides a robust and interpretable framework for spatio-temporal AGB prediction, offering practical value for planning of irrigation infrastructure, yield forecasting, and climate-adaptation planning.

\end{abstract}


\section{Introduction}
Accurately predicting the spatio-temporal variability of above-ground biomass (AGB) and associated water-stress dynamics is critical for food security, irrigation management, and climate-adaptation planning \cite{van2020crop}. Current approaches mainly rely on process-based crop models (PBCMs), such as DSSAT \cite{jones2003dssat}, APSIM  \cite{keating2003overview}, and LINTUL5 \cite{wolf2012user}, which represent crop development through mechanistic equations of radiation interception, phenological development, and stress responses. For example, the LINTUL5 model embedded in the SIMPLACE framework \cite{enders2023simplace} offers mechanistic insight into AGB formation with a water-stress coefficient. PBCMs are physiologically interpretable and can simulate latent variables such as leaf area index (LAI), photosynthetically active radiation (PAR), and radiation use efficiency (RUE). However, they typically require extensive calibration, site-specific parameterization, and detailed management information, which limits their scalability in large-scale or multi-environment applications. \par

Recently, data-driven machine learning models, such as random forests \cite{gumma2020agricultural}, multilayer perceptrons (MLPs) \cite{nivetha2024optimizing}, convolutional neural networks (CNNs) \cite{el2025review}, and, more recently, transformers \cite{xie2024recent}, have been widely applied to learn complex system behaviour at large scales \cite{cheng2024multimodal, dehghanisanij2022hybrid}. However, data-driven models generally require large training datasets, and their ``black-box'' nature limits their physiological interpretability and often leads to poor generalization across regions and seasons. For instance, Worrall \textit{et al.} show that data-driven machine-learning models can perform crop mapping and yield prediction, but they struggle to disentangle underlying crop-growth drivers or capture their changing importance throughout the growing season \cite{worrall2021domain}. Hybrid approaches that integrate domain knowledge into machine learning have proven effective for extrapolating beyond the distribution of the training dataset \cite{paudel2021machine}. However, many existing ``grey-box'' hybrid models insert domain knowledge only as extra features or ad-hoc adjustments, which significantly increases the number of trainable parameters \cite{ren2024based}. \par

Hybrid and process-informed methods attempt to combine the strengths of data-driven and process-based paradigms. Knowledge-guided machine learning \cite{liu2024knowledge} and process-informed neural networks \cite{hu2023ai} introduce biophysical constraints into deep learning models, improving generalization and interpretability. Recent agricultural applications demonstrate the potential of this approach, for example, Bai \textit{et al} \cite{bai2024evaluation} developed a hybrid model by incorporating the APSIM model outputs and the most informative growth stage-specific extreme climate indices (ECIs) into random forest and light gradient boosting machine, to evaluate impacts of climate extremes on wheat yields in the North China Plain. Their results showed that the hybrid model outperformed the process-based model in estimating wheat yield regardless of input variables. However, most existing hybrid modelling approaches rely on simplified crop processing, such as the spatial process of crop distribution \cite{kheir2023integrating}, and on estimation of crop variables \cite{droutsas2022integration}. These modelling approaches are rarely available at scale. \par

In this study, we present AgriPINN, a process-informed neural network that incorporates the LINTUL5 biomass-dynamics differential equation as a differentiable soft constraint during training. Unlike existing hybrid approaches, AgriPINN constrains learning with physiologically meaningful equations while using only AGB observations for supervision. This enables unsupervised recovery of latent physiological variables, including LAI, PAR, RUE, and water-stress factors, and links them to environmental and management factors—within a unified data-driven and process-based framework. Specifically, AgriPINN couples data supervision with differential-equation constraints to produce physiologically consistent spatio-temporal AGB predictions while simultaneously inferring crop physiological states (e.g. LAI, PAR, RUE). This dual-output design enables cross-validation between AGB predictions and inferred physiological states, supporting quantitative analysis of water-stress responses. AgriPINN is pretrained on 60 years of historical data across 397 regions of Germany and fine-tuned on three years of field experiments under controlled water conditions (i.e., shelter, rainfed, and irrigated). Our results show that AgriPINN outperforms both process-based and data-driven models in scalability, mechanistic consistency, and computational efficiency. The main contributions include:

\begin{itemize}
\item \textbf{Process-informed neural architecture.}  
      We propose AgriPINN, a novel process-informed neural network that embeds crop-growth differential equations as differentiable constraints during training. This design promotes physiologically consistent biomass predictions under water stress while keeping the neural architecture simple and training efficient.

\item \textbf{Unsupervised recovery of key latent physiological variables.}  
      Through constrained learning, AgriPINN infers unobserved variables, including LAI, APAR, RUE, and water-stress factors, without direct supervision. These latent estimates align with independent field observations, providing interpretability beyond conventional machine-learning approaches.

\item \textbf{Improved computational efficiency for large-scale applications.}
      AgriPINN eliminates the repeated parameter-calibration cycles required by process-based crop models and enables fast inference by using a compact neural architecture. As a result, AgriPINN achieves 8$\times$ faster inference than LINTUL5 while remaining physiologically consistent through differentiable growth constraints.
\end{itemize}

The remainder of this paper is organized as follows: Section~\ref{sec:related} describes the related work on data-driven, process-based and hybrid approaches for AGB estimation. The Section~\ref{sec:problem} describes the problem formulation, and the proposed AgriPINN, including the overall framework, backbone, and training strategies. Section~\ref{sec:material_experiment} outlines the materials and experimental methods, and Section~\ref{sec:results} presents the results. Section~\ref{sec:discussion} discusses the findings, and finally, conclusions are drawn in Section~\ref{sec:conclusion}.

\section{Related work}
\label{sec:related}

\subsection{Data-driven approaches for AGB estimation}
Machine learning has been widely applied to predict crop AGB and yield from environmental and crop-related variables. Early studies employed tree-based methods and shallow neural networks to capture nonlinear interactions between environmental variables and crop AGB. For example, Dhillon \textit{et al.} \cite{dhillon2023integrating} incorporated the Light Use Efficiency (LUE) into Random Forest (RF) models to improve biomass and yield prediction of winter wheat (WW) and oil seed rape (OSR) in the Free State of Bavaria, Germany, in 2019, demonstrating that data-driven models can reduce relative root mean square error (RRMSE). Halder \textit{et al} \cite{halder2025robust} compared XGBoost, CatBoost, Bidirectional LSTM (BiLSTM), and a Self-Attention-enhanced network for crop mapping cross three German states for 2021 and 2023. Their findings highlight that data-driven models demonstrated strong spatial transferability and adaptability, achieving high performance in new regions even with limited training data, and outperforming established benchmark datasets. \par

Recent advances in deep learning have expanded AGB estimation beyond traditional machine-learning methods, introducing diverse architectures—including CNNs, recurrent networks, and Transformers—that capture spatial patterns, temporal growth dynamics, and long-range dependencies in crop development. Convolutional neural networks (CNNs) have been widely applied to extract spatial features from remote-sensing imagery, while recurrent architectures such as long short-term memory (LSTM) networks are effective in modeling temporal growth trajectories. For example, Nakajima \textit{et al.} \cite{nakajima2023biomass} evaluated a CNN-based estimation method for rice AGB using digital images with 59 diverse cultivars. Their results indicate that CNNs can generalize across cultivar variability, suggesting their practicality for operational AGB monitoring. Gafurov \textit{et al.} \cite{gafurov2023advancing} showed that LSTM networks are capable of learning long-term temporal dependencies and can overcome limitations of traditional machine-learning techniques. Their results indicate that LSTM networks, although more computationally expensive, provide more accurate solutions for agricultural applications compared with MLP and RF algorithms. In addition, transformer-based architectures have demonstrated strong performance for yield forecasting by modeling long-range dependencies. Du \textit{et al.} \cite{du2025enhancing} developed a convolutional neural network (CNN)–Transformer architecture to estimate winter wheat biomass and yield by combining the local feature-extraction strengths of CNNs with the global information-capture abilities of Transformer networks using self-attention mechanisms. Jacome \textit{et al.} \cite{jacome2025agritransformer} proposed an AgriTransformer model, a Transformer-based architecture that enhances crop yield prediction by leveraging attention mechanisms for multimodal data fusion, incorporating tabular agricultural data and vegetation indices (VIs). However, despite strong predictive performance, these data-driven models typically require large labeled datasets, are prone to overfitting across regions, and lack physiological interpretability, limiting their utility for mechanistic understanding. \par

\subsection{Process-based crop models for the simulation of stress-driven AGB dynamics}

Process-based crop models (PBCMs) are widely used tools for analyzing interactions between crop growth, biomass production, and environmental drivers such as climate, soil properties, and management practices. These models simulate crop development by explicitly representing key physiological and biophysical processes that govern plant growth under varying environmental conditions \cite{van2003approaches}. Within broader agricultural and environmental assessments, PBCMs are often embedded in integrated modeling frameworks that link crop growth processes with climate, hydrology, and management systems to support systems-level analysis and decision-making \cite{laniak2013integrated}. PBCMs encompass a wide range of disciplines with expanding boundaries and increasing societal relevance, particularly in agricultural analysis \cite{nakayama2022impact, macpherson2020linking, couedel2024long}. In agricultural applications, PBCMs such as APSIM \cite{keating2003overview}, DSSAT \cite{jones2003dssat}, and WOFOST \cite{van1989wofost} simulate crop growth and aboveground biomass accumulation using weather, soil, and management inputs, while differing in their levels of process detail, structural complexity, and reliance on empirical parameterizations \cite{holzworth2014apsim}. These models represent physiological processes including photosynthesis, respiration, phenological development, and carbon allocation, which together determine the temporal evolution of aboveground biomass. Environmental stresses such as water or nitrogen limitation are typically incorporated through stress response functions that regulate carbon assimilation, growth rates, and biomass partitioning \cite{van2003approaches}. By mechanistically linking environmental drivers to crop physiological responses, PBCMs enable attribution of variations in aboveground biomass or yield to specific stress factors, supporting analysis of crop responses across contrasting environmental and management conditions. Multi-model studies have demonstrated that differences in process representation and parameterization can lead to substantial variability in simulated biomass and yield responses under stress, highlighting the importance of model structure in stress impact assessments \cite{asseng2013uncertainty}. For example, Kellner \textit{et al.} \cite{kellner2019response} coupled process-based hydrological and plant-growth models to simulate maize biomass under limited water supply. Ding \textit{et al.} \cite{ding2015modeling} extended a process-based multilayer model to explicitly simulate nonlinear biophysical processes within the crop canopy and achieved quantitative analysis of crop water use in the hydrologic cycle of the canopy. \par

PBCMs use mathematical representations of biophysical processes, combining mechanistic formulations with empirical parameterizations to simulate crop–environment interactions. These models serve critical functions, including elucidating system behaviors, testing theoretical constructs, and predicting responses to changes in external forcing or internal properties. Many PBCMs share common structural components, typically representing temporal dynamics through nonlinear differential or algebraic equations, most commonly ordinary differential equations (ODEs). For example, SIMPLACE (Scientific Impact Assessment and Modelling Platform for Advanced Crop and Ecosystem) \cite{enders2023simplace} is a modular simulation framework designed to capture plant–environment interactions with high precision. SIMPLACE integrates various process-based submodels and data sources for simulating agricultural systems, with a modular architecture that allows the combination of different components to address specific research questions. APSIM (Agricultural Production Systems Simulator) is a comprehensive modeling framework renowned for its holistic approach to simulating entire farming systems. Developed through collaborative efforts involving Australian and international researchers \cite{mccown1995apsim}, APSIM integrates a wide range of interrelated processes—including crop growth, soil biogeochemistry, water dynamics, and livestock management—within a unified and flexible platform. \par

The understanding of vegetation and crop growth processes underlying stress-driven biomass dynamics varies, with some processes well established and others relying on assumptions or empirical representations; this diversity in process representation contributes to structural and parametric uncertainty in model predictions \cite{asseng2013uncertainty}.
In the context of stress-driven aboveground biomass simulation, parameterizations are therefore required to represent processes that are either insufficiently understood or occur at spatial and temporal scales finer than those resolved by available climate or management data, which introduces structural and parametric uncertainty into model predictions. In agricultural impact assessments, PBCMs are commonly integrated with climate and management scenarios to support decision-making for farmers, resource managers, and policymakers \cite{ewert2015uncertainties, challinor2018improving}. \par

\subsection{Hybrid modeling approaches}
Hybrid modelling approaches aim to combine the complementary strengths of process-based crop models (PBCMs) and data-driven machine learning to improve prediction accuracy, mechanistic interpretability, and robustness under stress conditions. Machine-learning models contribute flexible, high-capacity function approximators capable of capturing complex nonlinear relationships that are difficult to specify explicitly within PBCMs. In contrast, PBCMs rely on a small set of interpretable, physiologically meaningful parameters, which enhances mechanistic clarity but limits flexibility and scalability compared with high-capacity deep learning models. \par

A widely used hybrid strategy is model ensembling, in which PBCM outputs are treated as additional features for a machine-learning model or combined with statistical predictions. For example, Shahhosseini \textit{et al.} \cite{shahhosseini2021coupling} created an ensemble model in which outputs from the process-based APSIM module were used as features in a machine-learning model, yielding improved maize-yield predictions. Another approach is data assimilation, where data-driven features are ingested into a process model to correct its state trajectory. For example, Tewes \textit{et al.} \cite{tewes2020methods} proposed an assimilated model in which Sentinel-2 LAI observations were incorporated to improve biomass simulations under water-stress conditions. Bai \textit{et al.} \cite{bai2024evaluation} incorporated the APSIM model outputs and the growth stage-specific extreme climate indices (ECIs) into two machine learning algorithms (random forest, RF and light gradient boosting machine, LGBM). Their results showed that the RF model outperformed the LGBM model in estimating wheat yield regardless of input variables, which reduces computational cost without significantly affecting model accuracy.\par

A more recent paradigm, inspired by the success of Physics-Informed Neural Networks (PINNs) in engineering, is to enforce governing differential equations as constraints within a neural network. This approach has been successfully evaluated in hydrology, climate modeling, and other dynamical-system applications. Recently, PINN-based studies have also appeared in crop modeling. For instance, Liu \textit{et al.} \cite{liu2024knowledge} explored knowledge-guided neural networks for agricultural systems, where the known monotonic effect of nutrient supply on yield was encoded into the network architecture. However, fully integrating the differential equations of a complex crop-growth model into deep learning is still in its early stages. Most prior works simplify the integration, and these hybrids often stop short of embedding the entire set of process equations due to challenges in differentiability and computational cost.

\section{The Proposed AgriPINN Framework}
\label{sec:problem}

This section formalizes the learning problem addressed by AgriPINN and explains why the proposed process-informed design of a deep-learning approach yields both interpretable predictions of ABG at large spatial scales and improved generalization. We consider the mapping from multi-source agro-environmental drivers $X(\mathbf{p},t)$ to the AGB trajectory $AGB(\mathbf{p},t)$ across large spatial domains and over the growing season. Rather than fitting $AGB$ with a purely data-driven regressor, we embed the core biomass dynamics of LINTUL5 into the network through a process residual, so that day-to-day biomass increments adhere to the process law up to a controlled tolerance. Subsection~3.1 specifies the process-informed formulation and defines the residual operator $\mathcal{L}$ that enters the training objective as a Tikhonov-style penalty. Building on this, Subsection~3.2 explains why the learned decomposition of biomass increments is semantically aligned with plant physiology. Subsection~3.3 then connects the process residual to statistical capacity control and dynamical robustness under covariate shift.

\subsection{Process-informed formulation to estimate crop AGB}

In this study, we aim to model the large-scale spatio-temporal dynamics of daily crop AGB across an entire growing season. Formally, let $X(\mathbf{p}, t)$ denote the daily input observations (i.e., climate, soil, management) for a given place $\mathbf{p}$ at time (day) $t$. We aim to predict crop AGB, denoted as $AGB(\mathbf{p}, t)$, through a function $F_\theta$, so that
\begin{equation}
AGB(\mathbf{p}, t) = F_\theta(X(\mathbf{p}, t)) + B,
\label{eq:initial}
\end{equation}
where $B$ is the systematic bias. To inform $F_\theta$, we employ the core biomass-dynamics ODE of the LINTUL5 model \cite{wolf2012user}, which is given by
\begin{equation}
AGB(t+1) = AGB(t) + RUE(t) \cdot I_{\text{int}}(t) \cdot F_W(t),
\label{eq:biomass1}
\end{equation}
where $RUE$ is the radiation-use efficiency, $I_{\text{int}}$ is the intercepted solar radiation, and $F_W \in [0,1]$ is a water-stress factor representing the effect of water stress on crop growth. This ODE is defined at a point and assumes field homogeneity, which limits its ability to capture spatial heterogeneity in large-scale settings. To address this, we extend the formulation by indexing all variables by location $\mathbf{p}$, obtaining a spatially indexed ODE system:
\begin{equation}
AGB(\mathbf{p}, t+1) = AGB(\mathbf{p}, t) + RUE(\mathbf{p}, t) \cdot I_{\text{int}}(\mathbf{p}, t) \cdot F_W(\mathbf{p}, t).
\label{eq:biomass}
\end{equation}

The intercepted radiation $I_{\text{int}}$ can be further related to the photosynthetically active radiation $PAR$ via the leaf area index ($LAI$), given by
\begin{equation}
I_{\text{int}}(\mathbf{p}, t) = PAR(\mathbf{p}, t) \cdot \big(1 - e^{-k \, LAI(\mathbf{p}, t)}\big),
\label{eq:par}
\end{equation}
where $k$ is the light-extinction coefficient. Combining Eq.~\ref{eq:par} with Eq.~\ref{eq:biomass} and introducing the discrete forward difference
\(
\Delta AGB(\mathbf{p}, t) = AGB(\mathbf{p}, t+1) - AGB(\mathbf{p}, t),
\)
we obtain
\begin{equation}
\Delta AGB(\mathbf{p}, t) = RUE(\mathbf{p}, t) \cdot PAR(\mathbf{p}, t) \cdot\big(1 - e^{-k\,LAI(\mathbf{p}, t)}\big)\cdot F_W(\mathbf{p}, t).
\label{eq:update}
\end{equation}

In our formulation, $AGB$ is the primary target variable, and the biophysical quantities $LAI$, $PAR$, $RUE$, and $F_W$ are treated as latent physiological variables $LATENT(\mathbf{p}, t)$. Eq.~\ref{eq:initial} can therefore be extended as
\begin{equation}
\{AGB(\mathbf{p}, t), LATENT(\mathbf{p}, t)\} = F_\theta(X(\mathbf{p}, t)) + B,
\label{eq:final}
\end{equation}
where
\(
LATENT(\mathbf{p}, t) = \{LAI(\mathbf{p}, t), PAR(\mathbf{p}, t), RUE(\mathbf{p}, t), F_W(\mathbf{p}, t)\}
\)
collects the latent biophysical states associated with AGB. 

To enforce adherence to the LINTUL5 dynamics in Eq.~\ref{eq:update}, we train the neural network $F_\theta$ with a process-informed objective that penalizes violations of Eq.~\ref{eq:update}. This objective comprises a standard empirical risk minimization (ERM) term $\mathcal{J}_{\text{data}}$ for the observed AGB, augmented by a regularization term $\mathcal{J}_{\text{proc}}$ that penalizes deviations from the governing process law.

The process-based residual operator $\mathcal{L}$ is defined as the deviation from the dynamic equation:
\begin{equation}
\mathcal{L}(\mathbf{p}, t) := \Delta AGB(\mathbf{p}, t) - \Phi(LATENT(\mathbf{p}, t)),
\label{eq:lossss}
\end{equation}
where $\Phi$ denotes the right-hand side of Eq.~\ref{eq:update}. The full training objective $\mathcal{J}(\theta)$ is a weighted sum of the data-fidelity term and the process-residual term:
\begin{equation}
\mathcal{J}(\theta) = \mathcal{J}_{\text{data}}(\theta) + \lambda \cdot \mathcal{J}_{\text{proc}}(\theta),
\label{eq:objective}
\end{equation}
where $\lambda > 0$ controls the trade-off between fitting the observed data and adhering to the LINTUL5 process dynamics. The data term $\mathcal{J}_{\text{data}}$ is the standard mean-squared error (MSE) on the available AGB observations $\mathcal{D}_d$:
\begin{equation}
\mathcal{J}_{\text{data}}(\theta) = \frac{1}{|\mathcal{D}_d|} \sum_{(\mathbf{p}, t) \in \mathcal{D}_d} \big( AGB(\mathbf{p}, t) - AGB_{\text{obs}}(\mathbf{p}, t) \big)^2.
\label{eq:objective11}
\end{equation}
The Tikhonov-style process penalty $\mathcal{J}_{\text{proc}}$ is constructed from the squared residual operator, defining the operator seminorm $\|\cdot\|_{\mathcal{L}}$ referenced in Claim~4. This term enforces the process constraint across a set of collocation points $\mathcal{D}_c$, typically the entire spatio-temporal domain of interest:
\begin{equation}
\mathcal{J}_{\text{proc}}(\theta) = \big\|(AGB, LATENT)\big\|_{\mathcal{L}}^2 = \frac{1}{|\mathcal{D}_c|} \sum_{(\mathbf{p}, t) \in \mathcal{D}_c} \big( \mathcal{L}(\mathbf{p}, t) \big)^2.
\label{eq:objective22}
\end{equation}

By minimizing $\mathcal{J}(\theta)$, AgriPINN learns a function $F_\theta$ that not only predicts AGB accurately but also constrains the learned day-to-day biomass increments to be process-consistent. As discussed in Subsections~3.2 and~3.3, this yields interpretable predictions and improved generalization, particularly under out-of-distribution (OOD) conditions.

\subsection{AgriPINN model architecture}

Having established in Section~3 the process-informed learning framework and its associated generalization and stability properties, we now turn to the practical realization. The architecture is designed to satisfy the structural requirements imposed by the biomass-dynamics ODE, enabling the network to predict both AGB and its latent physiological variables while maintaining physical consistency. These theoretical considerations motivate the form of the loss function, the structure of the latent space, and the use of a residual operator that enforces the process constraint. In this section, we develop an agricultural process-informed neural network (AgriPINN) to model $F_\theta$ with trainable parameters $\theta$ to estimate the spatio-temporal variability of crop $AGB(\mathbf{p}, t)$ and its physiological variables $LATENT(\mathbf{p}, t)$. Based on Eq.~\ref{eq:update} and Eq.~\ref{eq:final}, the optimization objective of AgriPINN is to seek $\theta$ that minimizes prediction error on observed data while satisfying the governing biomass-growth equation by penalizing the residuals as follows:
\begin{equation}
r(\mathbf{p}, t) = \Delta AGB(\mathbf{p}, t) - RUE(\mathbf{p}, t) \cdot PAR(\mathbf{p}, t) \cdot\big(1 - e^{-k\,LAI(\mathbf{p}, t)}\big)\cdot F_W(\mathbf{p}, t),
\label{eq:residual}
\end{equation}
where the increment $\Delta AGB(\mathbf{p}, t)$ represents a discrete forward difference and can be computed either from consecutive predictions $AGB(\mathbf{p}, t+1)-AGB(\mathbf{p}, t)$ or via automatic differentiation with respect to time. 

\begin{figure}[!t]
    \centering
    \includegraphics[width=5.5in]{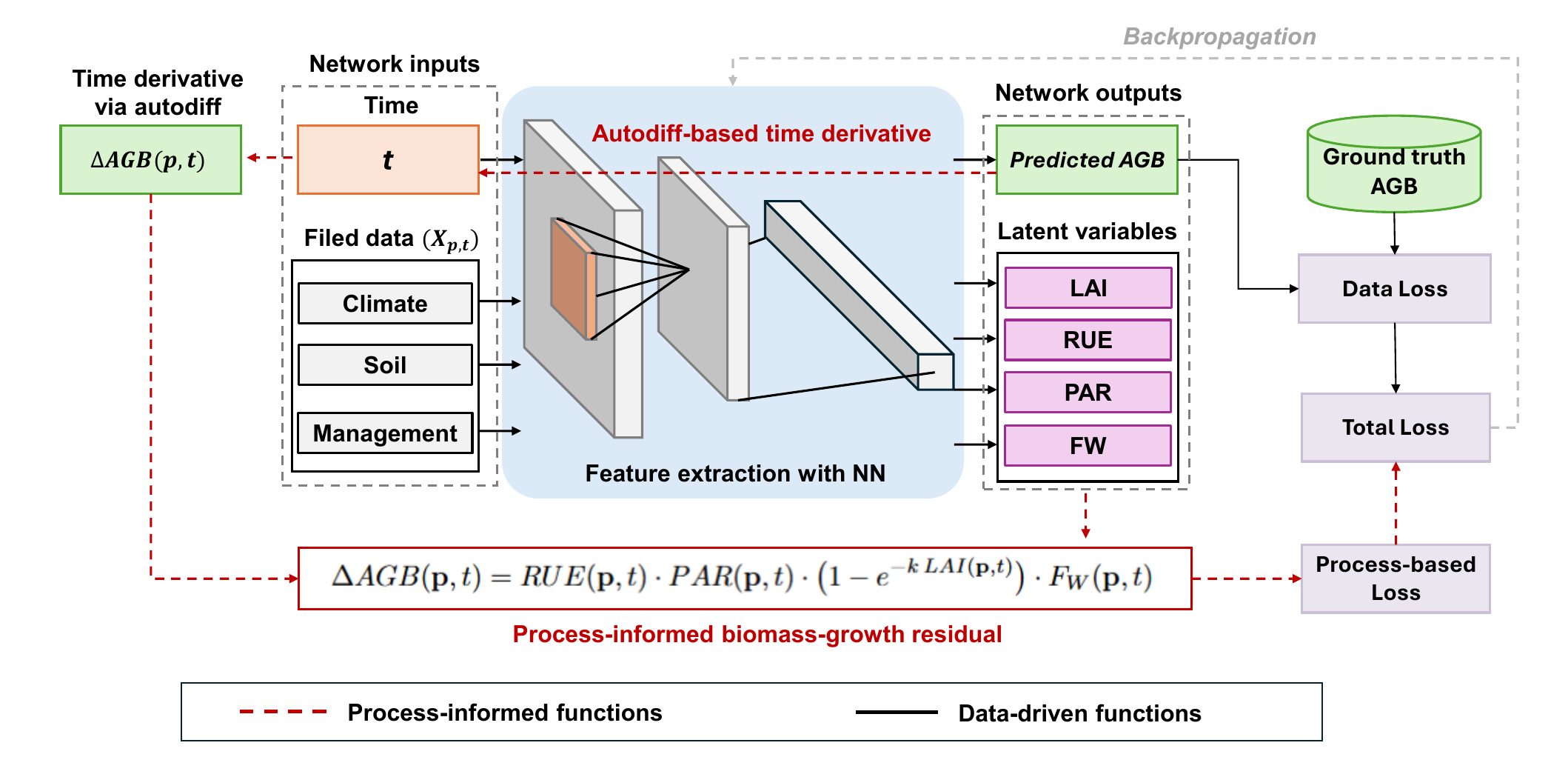}
    \caption{Workflow of the proposed AgriPINN model. The architecture integrates deep learning with process-based crop modeling by incorporating the LINTUL5 biomass-growth ODE as a soft constraint. The neural network predicts AGB together with latent physiological variables (LAI, RUE, PAR, $F_W$). These outputs are substituted into the biomass-growth equation to compute the process residual $r(\mathbf{p},t)$, where $\Delta AGB(\mathbf{p},t)$ is obtained as a time derivative using automatic differentiation. This residual defines the process-based loss, which is optimized jointly with the standard data loss to encourage biophysically consistent predictions across spatial locations and time steps.}
    \label{fig:1}
\end{figure}

To demonstrate the effectiveness of the AgriPINN design, we adopt a convolutional neural network (CNN) as the backbone. Figure~\ref{fig:1} shows the main framework of the proposed AgriPINN with a CNN backbone. Specifically, for a given location $\mathbf{p}$ and time $t$, the input $X(\mathbf{p}, t)$, including climate, soil, and management observations, is flattened as needed for the convolutional layers. Then, three convolutional layers with kernel sizes of $5\times5$ and ReLU activations are applied, followed by pooling layers to progressively reduce spatial dimensionality. This allows the model to extract deep features sensitive to biomass. After convolutional feature extraction, a fully connected layer is used to map the extracted features to $AGB(\mathbf{p}, t)$ and $LATENT(\mathbf{p}, t)$.  \par
    
For model regularization and training, the batch size is set to 8, updating the parameters after each forward pass. We employ stochastic gradient descent (SGD) with momentum as the optimizer, providing both accelerated convergence and resilience against local minima. The maximum number of training iterations is set to 1000, which balances training complexity with model accuracy, and the initial learning rate is 0.01. A dropout rate of 0.3 is maintained throughout training to enhance model generalization.  

The training objective can be written as a multi-term loss: 
\begin{equation}
L(\theta) = L_{\text{data}}(\theta) + \lambda L_{\text{phys}}(\theta),
\label{eq:loss_total}
\end{equation}
where $L_{\text{data}}$ is a data-fidelity term, which denotes the mean squared error (MSE) between predictions and ground truth, and $L_{\text{phys}}$ is a process-based soft constraint that embeds the process residuals in Eq.~\ref{eq:residual}. The hyperparameter $\lambda$ controls the trade-off between fitting the data and enforcing the process-based rule. Specifically, $L_{\text{data}}$ is defined as:
\begin{equation}
L_{\text{data}}(\theta) = \frac{1}{P} \frac{1}{T} \sum_{p=1}^{P} \sum_{t=1}^{T} \bigl(AGB(\mathbf{p}, t) - \hat{AGB}(\mathbf{p}, t)\bigr)^2, 
\label{eq:mse_biomass} 
\end{equation}
and $L_{\text{phys}}$ is formulated as:
\begin{equation}
L_{\text{phys}}(\theta) = \frac{1}{|\mathcal{B}|} \sum_{(\mathbf{p},t)\in \mathcal{B}} \bigl(r_\theta(\mathbf{p}, t)\bigr)^2,
\label{eq:physics_loss} 
\end{equation}
where $r_\theta(\mathbf{p}, t)$ is the process residual defined in Eq.~\ref{eq:residual}, and $\mathcal{B}$ denotes the (sub)samples used to approximate the process loss (e.g., a training batch or the full spatio-temporal domain). 

It is noteworthy that, based on the total loss in Eq.~\ref{eq:loss_total}, only the ground truth $\hat{AGB}(\mathbf{p}, t)$ is required during optimization; the prediction of physiological variables $LATENT(\mathbf{p}, t)$ is governed by the process-based loss in an unsupervised manner. Consequently, the outputs $LATENT(\mathbf{p}, t)$ can be used as self-consistency indicators for the predicted AGB. This design reduces the modeling cost associated with extensive data annotation while increasing model robustness and scalability.

\begin{algorithm}[t]
\caption{Training Procedure of the AgriPINN}
\label{alg:agripinn}
\small
\begin{algorithmic}[1]
\Require
  Spatio-temporal inputs $X(\mathbf{p},t)$ (climate, soil, management), observed biomass $\hat{AGB}(\mathbf{p},t)$; hyperparameters: process-weight $\lambda$, learning rate $\eta$, batch size $B$, maximum iterations $K_{\max}$.
\State \textbf{Define network output:} The network $F_\theta$ is defined such that its output for input $X(\mathbf{p}, t)$ is a tuple of the estimated AGB and the latent variables:
\[
\bigl( AGB_\theta(\mathbf{p}, t), LATENT_\theta(\mathbf{p}, t) \bigr) = F_\theta(X(\mathbf{p}, t)),
\]
where $LATENT_\theta = (RUE_\theta, PAR_\theta, LAI_\theta, F_{W_\theta})$.
\State \textbf{Initialize} network parameters $\theta$ of $F_\theta$; set the optimizer to stochastic gradient descent (SGD) with momentum and learning rate $\eta$.
\State \textbf{Training objective:}
\begin{equation*}
L(\theta) = L_{\mathrm{data}}(\theta) + \lambda L_{\mathrm{phys}}(\theta).
\end{equation*}
\State \textbf{Data loss (Empirical Risk Minimization):}
\begin{equation*}
L_{\mathrm{data}}(\theta)
=
\frac{1}{P\,T}
\sum_{p=1}^{P} \sum_{t=1}^{T}
\bigl( AGB_\theta(\mathbf{p}, t) - \hat{AGB}(\mathbf{p}, t) \bigr)^2.
\end{equation*}
\State \textbf{Process loss (Tikhonov penalty):}
\begin{equation*}
L_{\mathrm{phys}}(\theta)
=
\frac{1}{|\mathcal{B}|}
\sum_{(\mathbf{p},t)\in\mathcal{B}}
\bigl(r_\theta(\mathbf{p}, t)\bigr)^2,
\end{equation*}
where $r_\theta(\mathbf{p}, t)$ is the process residual (LINTUL5 dynamics violation):
\begin{equation*}
r_\theta(\mathbf{p}, t)
=
\Delta AGB_\theta(\mathbf{p}, t)
-
RUE_\theta(\mathbf{p}, t)\cdot PAR_\theta(\mathbf{p}, t)\cdot
\bigl(1 - e^{-k\,LAI_\theta(\mathbf{p}, t)}\bigr)\cdot F_{W_\theta}(\mathbf{p}, t),
\end{equation*}
and $\Delta AGB_\theta(\mathbf{p}, t)$ is the finite difference or temporal derivative of $AGB_\theta(\mathbf{p},t)$.
\For{$k = 1$ \textbf{to} $K_{\max}$}
  \State Sample a mini-batch $\mathcal{B} = \{(\mathbf{p}_i,t_i)\}_{i=1}^{B}$ from the training set.
  \State Construct input tensors $\{X(\mathbf{p}_i,t_i)\}_{i=1}^{B}$ and observed AGB $\{\hat{AGB}(\mathbf{p}_i,t_i)\}_{i=1}^{B}$.
  \State \textbf{Forward pass to obtain outputs:}
    \Statex \hspace{1.2em} $\bigl( AGB_\theta(\mathbf{p}_i, t_i), LATENT_\theta(\mathbf{p}_i, t_i) \bigr) \leftarrow F_\theta(X(\mathbf{p}_i,t_i))$ for all $i\in\mathcal{B}$.
  \State Compute $\Delta AGB_\theta(\mathbf{p}_i,t_i)$ and the residual $r_\theta(\mathbf{p}_i,t_i)$ on the batch $\mathcal{B}$.
  \State Evaluate $L_{\mathrm{data}}(\theta)$ and $L_{\mathrm{phys}}(\theta)$, and the total loss $L(\theta)$.
  \State \textbf{Backward pass:} compute gradients $\nabla_\theta L(\theta)$ by backpropagation.
  \State \textbf{Parameter update:} update $\theta \leftarrow \theta - \eta\,\mathrm{SGD\_momentum}\bigl(\nabla_\theta L(\theta)\bigr)$.
\EndFor
\State \textbf{Return} the trained parameters $\theta^\star$ and the final network output:
\[
\bigl( AGB_{\theta^\star}(\mathbf{p}, t), LATENT_{\theta^\star}(\mathbf{p}, t) \bigr).
\]
\end{algorithmic}
\end{algorithm}

\subsection{Process-informed semantics and identifiable elasticities}

Conventional black-box models can achieve high in-sample accuracy by exploiting spurious correlations or over-parameterized latent factors that lack physiological meaning. This often leads to poor out-of-distribution (OOD) generalization and undermines counterfactual or scenario-based analysis. By contrast, AgriPINN constrains day-to-day biomass updates to follow the LINTUL5 biomass-growth law and parameterizes crop responses through latent variables that are biophysically bounded and interpretable. In this subsection, we explain in theory why the proposed AgriPINN confers both interpretability and predictive strength.

\paragraph{Claim 1 (Semantic alignment).}
Once the LINTUL5 biomass-dynamics ODE system of Eq.~\ref{eq:biomass}--\ref{eq:update} is well-posed, any limit point $AGB(\mathbf{p}, t), LATENT(\mathbf{p}, t)$ satisfying $\mathcal{L}[AGB, LATENT](\mathbf{p}, t)=0$ almost everywhere also satisfies
\begin{equation}
AGB(\mathbf{p}, t{+}1)-AGB(\mathbf{p}, t)=\Phi(LATENT(\mathbf{p}, t)).
\label{eq:c1}
\end{equation}
Therefore, the learned day-to-day biomass increment is exactly decomposed into interpretable, unit-consistent factors $\mathrm{RUE}$, $\mathrm{PAR}$, $\mathrm{LAI}$, and $\mathrm{F_W}$, which inherit clear physical meaning.

\paragraph{Claim 2 (Physically grounded local elasticities).}
Under Eq.~\ref{eq:c1} and smooth $\Phi$,
\begin{equation}
\frac{\partial\,\Delta \mathrm{AGB}}{\partial\,\log \mathrm{PAR}}
=\Phi(LATENT)\quad\text{and}\quad
\frac{\partial\,\Delta \mathrm{AGB}}{\partial\,\mathrm{LAI}}
= \mathrm{RUE}\cdot \mathrm{PAR}\cdot \mathrm{F_W}\cdot k\,e^{-k\,\mathrm{LAI}},
\label{eq:c2}
\end{equation}
so the elasticities and signs computed from the learned model match the analytical elasticities of the LINTUL5 formulation. Explanations produced by AgriPINN correspond to mechanistic derivatives rather than arbitrary saliency scores.

\paragraph{Claim 3 (Latent identifiability up to physical symmetries).}
Suppose $\mathrm{RUE}$ is bounded in $[\underline r,\overline r]$, $\mathrm{F_W}\in[0,1]$, $\mathrm{LAI}\ge0$, and $\mathrm{PAR}$ is known/measured on $\mathcal{D}_v$. Then any two latent sets $LATENT, LATENT'$ that produce the same increments $\Phi(LATENT)=\Phi(LATENT')$ must coincide almost everywhere, except for known degeneracies (e.g., simultaneous proportional changes ruled out by the bounds and partial supervision). Hence $F_\theta$ learns biophysically unique components.

Together, these claims provide a mechanistic justification for AgriPINN’s improved out-of-distribution robustness, reduced variance across seasons and sites, and credible counterfactual behavior under alternative management or weather interventions. Detailed proofs of these claims are provided in the Supplementary Materials sections 1-3.

\subsection{Generalization with smaller effective hypothesis class and out-of-distribution (OOD) stability}

The claims in this subsection formalize why a process-informed objective improves test performance, particularly under out-of-distribution (OOD) conditions. Pure empirical risk minimization (ERM) over expressive neural classes can fit training data while exploiting spurious correlations, yielding large hypothesis complexity and brittle behavior under covariate shift. By contrast, the AgriPINN objective augments ERM with an operator-based Tikhonov penalty that contracts the feasible set to trajectories with small process residual and valid initial conditions, together with physically plausible temporal evolution. This defines a truncated hypothesis class $\mathcal{F}_\varepsilon$.

\paragraph{Claim 4 (Capacity control).}
Let $\mathcal{F}_\varepsilon$ denote the truncated hypothesis class
\[
\mathcal{F}_{\varepsilon}
:=
\big\{(AGB, LATENT)\in\mathcal{F}:\ \| (AGB, LATENT) \|_{\mathcal{L}}\le\varepsilon \ \text{and valid initial conditions hold}\big\},
\]
where $\mathcal{F}$ is the function class realizable by the neural network $F_\theta$ and $\|\cdot\|_{\mathcal{L}}$ is the operator seminorm. Assume that the network $F_\theta$ has bounded weights and employs Lipschitz activation functions. Then there exists a constant $C>0$ such that the empirical Rademacher complexity satisfies
\begin{equation}
\mathfrak{R}_n\!\left(\{AGB:(AGB, LATENT)\in\mathcal{F}_{\varepsilon}\}\right)
\;\le\; C\,\varepsilon \;+\; \mathfrak{R}_n(\text{IC-feasible constants}) .
\label{eq:cc11}
\tag{11}
\end{equation}
In particular, as the residual tolerance $\varepsilon\downarrow 0$, the effective capacity of the process-consistent hypothesis class shrinks linearly in $\varepsilon$.

Claim~4 establishes that this truncation reduces the empirical Rademacher complexity at a rate linear in the residual tolerance \(\varepsilon\), which directly tightens standard generalization bounds and explains improved i.i.d.\ performance at fixed sample size. 

\paragraph{Corollary 1 (Generalization gap).}
Let $\ell$ be an $L$-Lipschitz loss function in its first argument, and let $\mathcal{R}(AGB)$ and $\widehat{\mathcal{R}}_n(AGB)$ denote the population and empirical risks, respectively. Then the expected generalization gap of the process-informed model satisfies
\begin{equation}
\mathbb{E}\big[\mathcal{R}(AGB_\theta)-\widehat{\mathcal{R}}_n(AGB_\theta)\big]
< L\,\mathfrak{R}_n\!\left(\{AGB:(AGB, LATENT)\in\mathcal{F}_{\varepsilon}\}\right)
+\mathcal{O}\!\big(n^{-1/2}\big),
\tag{12}
\end{equation}
and, by Claim~4, this bound improves linearly as $\varepsilon\downarrow 0$, in contrast to pure ERM over the full network class $\mathcal{F}$.

\paragraph{Claim 5 (Distribution-shift robustness).}
Assume that the test regime $\mathbb{P}_{\mathrm{test}}(X)$ differs from the training regime $\mathbb{P}_{\mathrm{train}}(X)$, but that under both regimes the true process $AGB^\star$ satisfies the biomass balance with latent variables $LATENT^\star$. Let $AGB_\theta$ denote the learned model. Suppose that on the training distribution the prediction error of $AGB_\theta$ is bounded by $\epsilon_{\mathrm{fit}}$ in an appropriate risk or norm, and that the operator seminorm of the learned pair satisfies
\[
\epsilon_{\mathrm{proc}} = \big\|(AGB_\theta,LATENT_\theta)\big\|_{\mathcal{L}}.
\]
If $\Phi$ is Lipschitz on the agronomic box and the discrete recurrence is stable with stability constant $C_{\mathrm{stab}}>0$, then for any test point $(\mathbf{p},t)$ drawn from $\mathbb{P}_{\mathrm{test}}(X)$,
\begin{equation}
|AGB_\theta(\mathbf{p},t)-AGB^\star(\mathbf{p},t)| \;\le\; C_{\mathrm{stab}}\,\big(\epsilon_{\mathrm{fit}}+\epsilon_{\mathrm{proc}}\big).
\label{eq:cc13}
\tag{13}
\end{equation}
In particular, the contribution of the process residual to the error bound is independent of the shift between $\mathbb{P}_{\mathrm{train}}$ and $\mathbb{P}_{\mathrm{test}}$, so the model inherits a form of out-of-distribution stability from the biomass dynamics.

Claim~5 provides a dynamical stability argument that complements the capacity-control result in Claim~4. When the physiological law governing biomass accumulation remains unchanged across regimes, the error on shifted test inputs is bounded by the sum of the in-distribution fit error and the process residual, scaled by a distribution-agnostic stability constant. Thus, AgriPINN’s prediction error under covariate shift cannot grow arbitrarily—it is constrained by the stability of the biomass-growth recurrence. Together, these results explain AgriPINN’s observed robustness under leave-year and leave-site evaluation protocols and its reduced variance across seasons. The model generalizes better because it searches a smaller, process-consistent hypothesis class, and it remains stable because its errors are governed by the law of biomass accumulation rather than by unconstrained correlations in the data. Detailed proofs of these claims are provided in the Supplementary Materials section 4-6.

\section{Material and Experimental Design}
\label{sec:material_experiment}

\subsection{Experimental overview}
In this section, we describe the datasets, baseline models, evaluation metrics, and implementation details used to validate the proposed AgriPINN framework. The experimental design is aligned with the four analytical components presented in Section~\ref{sec:results}, covering pretrained model evaluation, fine-tuned temporal variability simulation, large-scale spatial variability retrieval, and ablation studies on alternative backbones. The experiments are structured to ensure a rigorous and fair comparison between hybrid modeling, process-based simulation, and data-driven baselines. Accordingly, this section provides details on:

\begin{itemize}
    \item the pretraining dataset compiled from 65 years of crop aboveground biomass (AGB) simulations, generated with daily outputs from SIMPLACE for winter wheat and maize, used to build the noise-free supervisory signals that stabilise generalizable representations evaluated in Section~\ref{sec:pretrained};

    \item three years of real-world winter wheat and maize experiments under diverse water-stress regimes (rainout shelter, rainfed, and irrigated), enabling the pretrained representations to adapt to practical physiological conditions and the domain-specific biases induced by water stress \cite{nguyen2022responses};

    \item the comparative baseline models, including the process-based LINTUL5 simulator and three representative deep learning models (ConvLSTM-ViT, SLTF, and CNN-Transformer), which are compared against AgriPINN in terms of accuracy, stability, and computational efficiency;

    \item the evaluation metrics used throughout the study, including RMSE, $R^2$, CC, and RMSPE, as well as the statistical significance assessment using paired Wilcoxon signed-rank test, supporting both overall performance comparisons and variability analyses;

    \item the implementation details ensuring reproducibility and computational fairness, including consistent training budgets across models, unified data preprocessing pipelines, and runtime comparisons covering pretraining, fine-tuning, and inference phases.
\end{itemize}

This experimental setup provides a unified foundation for the four categories of analyses reported in Section~\ref{sec:results}: (i) pretrained performance on historical large-scale data, (ii) temporal AGB simulation under controlled filed and under water-stress conditions, (iii) spatial AGB retrieval across Germany using super-resolved ERA5 inputs, and (iv) ablation studies demonstrating the generality and efficiency of the proposed hybrid modeling approach across multiple neural network backbones.

\subsection{Datasets}
To rigorously evaluate the proposed AgriPINN framework, we leverage two complementary real-world datasets that capture diverse agro-environmental conditions and varying levels of water stress.

\textbf{1) Training dataset} \par

The pretraining dataset is derived from 65 years of regional-scale crop aboveground biomass (AGB) simulations cross germany, generated with daily outputs from SIMPLACE for winter wheat and maize \cite{nguyen2022responses}, together with associated environmental and management factors. The dataset spans 397 regions, classified at the EU NUTS-3 level, over the period 1951–2015. Key variables include: 
	\begin{itemize}
    \item \textbf{Climatic factors}: daily solar radiation ($I_0$, MJ/m²), minimum/maximum temperature (°C), and precipitation;
	\item \textbf{Soil characteristics}: soil type, organic matter content, and texture;
	\item \textbf{Management practices}: fertilizer application rates and irrigation schedules.
    \end{itemize}

The dataset contains 24,375 samples, each representing a unique combination of environmental and management conditions leading to an observed biomass outcome. Environmental variables are normalized for scale consistency, and missing values are imputed using mean substitution. The dataset is partitioned into $80\%$ training and $20\%$ validation sets using a stratified random split that preserves the distribution of crop types and years, enabling robust evaluation across diverse agro-environmental settings. \par


\textbf{2) Fine-tuning dataset} \par
The controlled water-treatment experiment was conducted at the TERENO Eifel/Lower Rhine Observatory near Selhausen, Germany ($50^{\circ}52^{\prime}N$, $6^{\circ}27^{\prime}E$). The experiment spanned three growing seasons (2016–2018) and examined the responses of wheat (2016) and maize (2017–2018) under three water regimes: sheltered, rainfed, and irrigated conditions. Wheat and maize cultivars were selected from varieties commonly grown in nearby farmers’ fields to ensure agronomic relevance. Additional experimental details are provided in \cite{nguyen2024multi}.

\textbf{3) Spatial variability dataset: high-resolution environmental inputs} \par
For national-scale spatial AGB retrieval, we integrate:
\begin{itemize}
    \item \textbf{ERA5 Post-Processed Daily Climate Statistics} at $0.25^\circ$ resolution, used as the primary climate forcing inputs;
    \item \textbf{MODIS MOD09Q1 NDVI product} for super-resolution refinement, enabling the upscaling of ERA5 data to 250\,m spatial resolution;
    \item \textbf{Ground-truth AGB}: regional AGB estimates derived from German yield statistics \cite{duden2024german}, converted to biomass via a harvest index of 0.55 as recommended by \cite{rose2017yield}.
\end{itemize}
These datasets enable the regression analysis in Fig.~\ref{fig:6_1} and the spatial AGB maps in Fig.~\ref{fig:6_2}, as well as the water-stress reconstruction in Fig.~\ref{fig:6_3}.

\textbf{4) Process-based simulation dataset (LINTUL5 baseline)} \par
The LINTUL5 process-based model is used as a baseline for both temporal (Section~\ref{sec:results}) and spatial analyses. To ensure fair comparison with AgriPINN, LINTUL5 is driven by the same climate, soil, and management forcings used in both the pretraining and spatial variability experiments. Simulated outputs include biomass, LAI, PAR, RUE, and $F_w$, serving as process-based references.

\medskip
Across all datasets, missing values are handled using standardized preprocessing, environmental variables are normalized, and inputs are synchronised temporally to maintain physical consistency across climate, soil, and management forcings. Together, these datasets support the comprehensive evaluation of AgriPINN across pretraining, temporal and spatial variability analyses, and hybrid modeling ablations.
\subsection{Comparative Methods}
\textbf{1) Data-driven models}\par
To verify the improved accuracy and generalization of the proposed hybrid model, three state-of-the-art (SOTA) data-driven models, including CNN-Transformer \cite{du2025enhancing}, SSA-LSTM-transformer (SLTF) \cite{guo2024novel}, and ConvLSTM-ViT \cite{mirhoseininejad2024convlstm} are used as baseline data-driven methods for comparison. All baseline models were implemented strictly according to their original configurations and hyperparameter settings as reported in the respective publications to ensure fairness and reproducibility. To maintain consistency across experiments, the same dataset partitioning ($80\%$ training, $12\%$ validation), input preprocessing, and optimization settings were uniformly applied to all models. Each model was trained for up to 200 epochs using the same computational environment (three NVIDIA 2080 Ti GPU, PyTorch 2.1 framework) to eliminate hardware-induced variability.

\textbf{2) Process-based models}\par
To evaluate the proposed efficiency and accuracy of the proposed model, we employed the generic LINTUL5 model implemented in the modeling framework SIMPLACE to simulate daily biomass development. Soil water balance was simulated using SlimWater, where the daily change in soil water content in a multiple layered soil profile is based on the volumes of crop water uptake, soil evaporation, surface run-off, and seepage below the root zone \cite{tewes2020methods}. Root growth was simulated using the SimComponent SlimRoots, where the daily increase in biomass of seminal and lateral roots depends on the input of assimilated biomass from the shoot \cite{seidel2022simulating}. We assumed no occurrence of disease stress and optimal nutrient supply at all times, as all fields were conventionally managed in accordance with local recommendations. Thus, water stress was the only growth-limiting factor considered in the model. For all crop specific variables beyond phenology, we used the generic values (i.e., no calibration of cultivar specifics). Daily meteorological data needed in the simulation model was taken from the data provided by the weather stations.

\subsection{Evaluation metrics}

To quantify the estimation performance of the proposed framework, Root Mean Square Error (RMSE) is employed as the primary metric. RMSE serves as the response variable, indicating the discrepancies in magnitude between the predicted crop biomass and the ground truth values. It is calculated as:\par

\begin{equation}
\text{RMSE} = \sqrt{\frac{1}{T} \sum_{t=1}^{T} (y_t - \hat{y}_t)^2}
\label{eq:rmse}
\end{equation}

where $y_t$ represents the actual crop AGB at time step $t$, and $\hat{y}_t$ denotes the corresponding predicted AGB. Here, $T$ signifies the total number of samples.\par

Pearson’s Correlation Coefficient (CC) is also utilized as an additional metric to assess the linear relationship between the predicted and actual crop AGB. It is calculated by:

\begin{equation}
\text{CC} = \frac{\sum_{t=1}^{T} (y_t - \bar{y})(\hat{y}_t - \bar{\hat{y}})}{\sqrt{\sum_{t=1}^{T} (y_t - \bar{y})^2 \sum_{t=1}^{T} (\hat{y}_t - \bar{\hat{y}})^2}}
\label{eq:cc}
\end{equation}

where $\bar{y}$ and $\bar{\hat{y}}$ denote the mean values of the ground truth and predicted crop biomasss, respectively.\par

Together, RMSE and CC provide a comprehensive assessment of model accuracy and reliability. RMSE quantifies average prediction error in the same units as AGB, while CC measures the strength and direction of the linear relationship between predictions and observations. These complementary metrics ensure that the framework meets the performance requirements for practical agricultural applications.\par

Furthermore, the statistical significance of the performance differences between the proposed framework and SOTA methods is determined using paired Wilcoxon signed-rank test, following the procedure described in \cite{waqar2025stacking}.

\subsection{Implementation details}

All experiments were implemented in PyTorch 2.1 and executed on a high-performance workstation equipped with three NVIDIA RTX 2080 Ti GPUs (12 GB memory each). Training was performed using mixed-precision computation to improve efficiency while maintaining numerical stability.
For the optimization procedure, we adopted the Adam optimizer with an initial learning rate of 0.001, coupled with a cosine annealing learning-rate scheduler that gradually reduces the learning rate over the course of training. The batch size was set to 128, and the maximum number of training iterations was 1,000, which provided a balance between computational cost and convergence stability. A dropout rate of 0.3 was applied to fully-connected layers for regularization, consistent with the training configuration described earlier.
The process-informed loss weight $\lambda$ was tuned via grid search from 0.05 to 1 with a step of 0.05 using the validation set, ensuring an appropriate balance between data fidelity and process consistency. All input features were standardized to zero mean and unit variance, and temporal windows were padded to accommodate automatic differentiation of the AGB increments.
To promote reproducibility, all random seeds were fixed to 42, and deterministic computation modes were enabled where possible without significantly affecting runtime. Training logs, model checkpoints, hyperparameter settings, and evaluation scripts were stored. \par

\section{Experimental results}
\label{sec:results}



\subsection{Model performance evaluation}
\label{sec:pretrained}

The pretrained AgriPINN and state-of-the-art (SOTA) methods (ConvLSTM-ViT, SLTF, CNN-Transformer) are evaluated on 15-year winter wheat and maize biomass datasets. Table~\ref{tab:1_0} reports the prediction performance of the proposed model and the SOTA baselines. The results indicate that AgriPINN achieves lower RMSE and higher CC than the competing methods. This improvement reflects the benefit of combining data supervision with process-based constraints, which helps reduce overfitting and promotes physiologically consistent predictions beyond what purely data-driven CNN-based architectures can achieve. 

To quantify performance differences, we conducted pairwise statistical comparisons between AgriPINN and each baseline model separately for winter wheat and maize. Because all models were evaluated on identical test samples, statistical significance was assessed using a paired Wilcoxon signed-rank test applied to per-sample MAE values.
Table ~\ref{tab:biomass_performance} summarises the average RMSE and MAE results for winter wheat and maize biomass prediction. AgriPINN achieves the lowest RMSE for both crops, indicating improved predictive accuracy compared with all benchmark models. Improvements over strong baselines, including CNN-Transformer and SLTF, are statistically significant, as indicated by the corresponding p-values reported in Table ~\ref{tab:biomass_performance}.
In addition, AgriPINN consistently yields lower MAE values across both crops, indicating reduced average prediction error. These performance gains are achieved despite AgriPINN’s comparatively lightweight architecture, allowing it to match or surpass more parameter-intensive models such as CNN-Transformer and ConvLSTM-ViT.

To quantify performance differences, we conducted pairwise statistical comparisons between AgriPINN and each baseline model separately for winter wheat and maize. Because all models were evaluated on identical test samples, statistical significance was assessed using a paired Wilcoxon signed-rank test applied to per-sample MAE values.

Table~\ref{tab:biomass_performance} summarises the average RMSE and MAE results for winter wheat and maize biomass prediction. AgriPINN achieves the lowest RMSE for both crops, indicating improved predictive accuracy compared with all benchmark models. The improvements over strong baselines, including CNN-Transformer and SLTF, are statistically significant ($p<0.01$ and $p<0.05$, respectively), as indicated in Table ~\ref{tab:biomass_performance}. In addition, AgriPINN consistently yields lower MAE values across both crops, suggesting reduced average prediction error. These performance gains are achieved despite AgriPINN’s lightweight architecture, enabling it to match or surpass more parameter-intensive models such as CNN-Transformer and ConvLSTM-ViT.

\begin{figure}[]   
    \centering  
    \includegraphics[width=5.5in]{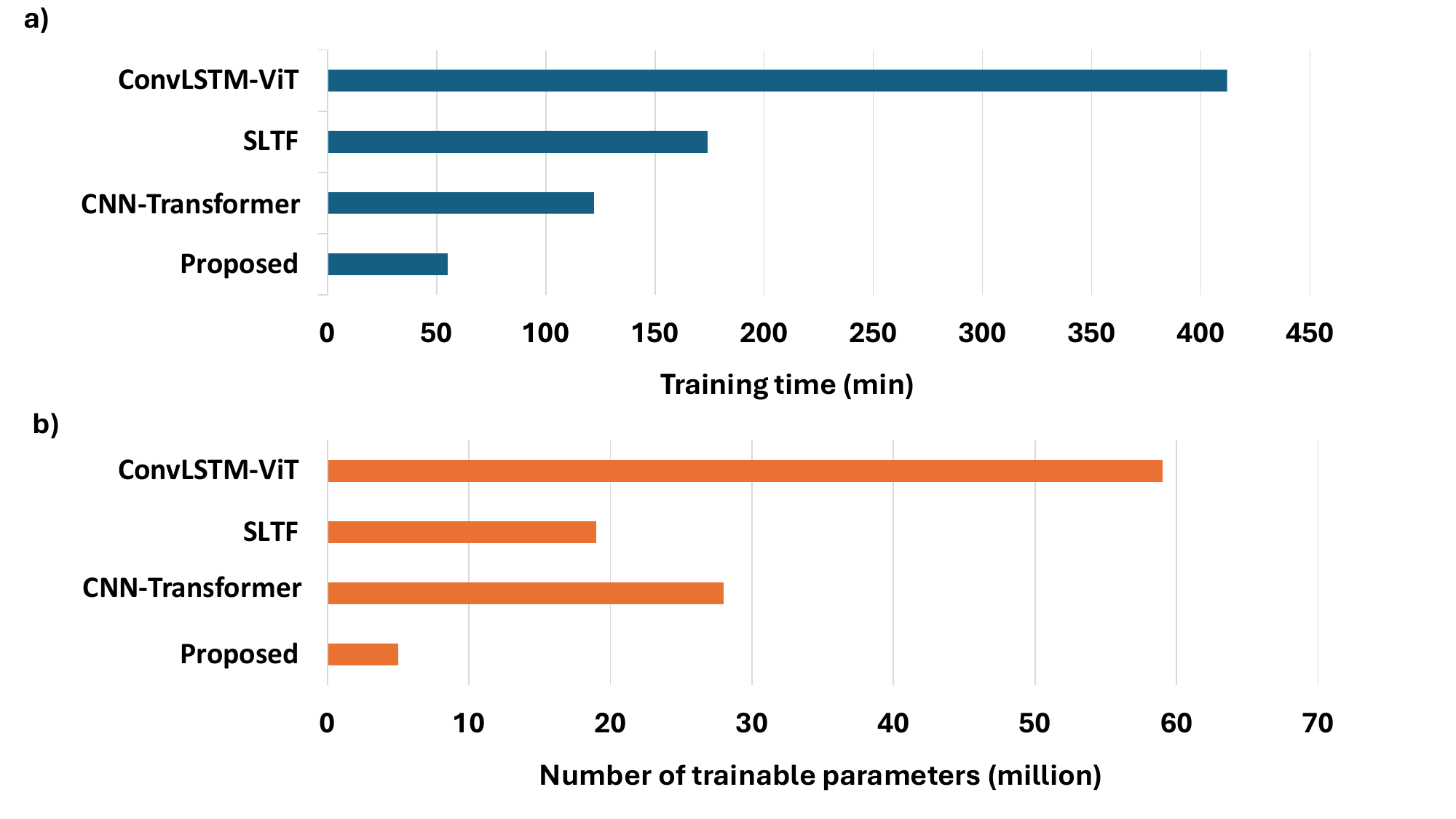}  
    \caption{Comparison of (a) training time and (b) model size for AgriPINN and the baseline data-driven architectures. Training time (in minutes) is measured for a full pretraining run on a single NVIDIA RTX 2080 Ti GPU under identical batch size, optimizer, and learning-rate settings. Model size is reported as the number of trainable parameters (millions). AgriPINN requires substantially less training time and has a significantly smaller parameter count than the deeper transformer-based models, reflecting its computational efficiency despite achieving higher predictive accuracy.} 
    \label{fig:11_runtime}  
\end{figure}

Figure~\ref{fig:11_runtime} presents the computational efficiency of AgriPINN compared with the baseline models. The proposed model has the shortest training time and the smallest parameter count among all evaluated methods. These findings motivate further incorporation of process-based agronomic knowledge into deep neural networks to enhance predictive performance.

\begin{table}[]
\caption{RMSE (t/ha) and CC of pretrained models for winter wheat and maize biomass prediction. Results compare three deep learning models (CNN-Transformer, SLTF, ConvLSTM-ViT) with the proposed AgriPINN model. The upper block reports RMSE (t/ha), and the lower block reports CC; lower RMSE and higher CC indicate better predictive accuracy.}
\label{tab:1_0}
\centering
\resizebox{4.2in}{!}{
\begin{tabular}{lcccc}
\toprule
 & \multicolumn{4}{c}{RMSE (t/ha)} \\ 
\cmidrule(lr){2-5}
Crop         & CNN-Transformer & SLTF        & ConvLSTM-ViT & Proposed AgriPINN \\ 
\midrule
Winter wheat & 2.08            & 1.43        & 1.62         & 1.42              \\
Maize        & 2.23            & 2.06        & 1.93         & 1.81              \\ 
\midrule
 & \multicolumn{4}{c}{CC} \\ 
\cmidrule(lr){2-5}
        & CNN-Transformer & SLTF        & ConvLSTM-ViT & Proposed AgriPINN \\ \midrule

Winter wheat & 0.78            & 0.88        & 0.82         & 0.94              \\
Maize        & 0.72            & 0.81        & 0.87         & 0.92              \\ 
\bottomrule
\end{tabular}
}
\end{table}


\begin{table}[h]
\caption{Biomass prediction performance for winter wheat and maize. Root mean square error (RMSE) and mean absolute error (MAE) are reported in tonnes per hectare (t/ha), with lower values indicating better performance. Statistical significance is assessed using a paired Wilcoxon signed-rank test, comparing each baseline model against the Proposed (AgriPINN) model based on per-sample MAE values, conducted separately for winter wheat and maize. The Proposed model serves as the reference and therefore has no associated p-value.}
\label{tab:biomass_performance}
\centering
\resizebox{\textwidth}{!}{
\begin{tabular}{lcccccc}
\toprule
& \multicolumn{3}{c}{Winter wheat} & \multicolumn{3}{c}{Maize} \\
\cmidrule(r){2-4} \cmidrule(l){5-7}
Method & RMSE (t/ha) & MAE (t/ha) & p-value (Proposed vs. SOTA)& RMSE (t/ha) & MAE (t/ha) & p-value (Proposed vs. SOTA) \\ 
\midrule
Proposed        & \textbf{1.42} & \textbf{0.21} & — & \textbf{1.79} & \textbf{0.25} & — \\
CNN-Transformer & 2.02 & 0.34 & $0.042$ & 2.32 & 0.38 & $0.039$ \\
SLTF            & 1.56 & 0.22 & $0.007$ & 2.08 & 0.29 & $0.021$ \\
ConvLSTM-ViT    & 1.65 & 0.33 & $0.003$ & 1.83 & 0.37 & $0.009$ \\ 
\bottomrule
\end{tabular}
}
\end{table}


\subsection{Temporal variability of crop AGB under different water-stress conditions}

To evaluate temporal AGB responses to water stress, we fine-tune the pretrained models using winter wheat and maize datasets collected under controlled water-stress treatments (sheltered, rainfed, and irrigated). Table~\ref{tab:1} summarizes the temporal prediction accuracy for all models across the three water-stress treatments. AgriPINN achieves the best overall performance, with an average RMSE of 2.01, $R^2 = 0.93$, and CC $= 0.92$ across all treatments. Compared with the pretrained evaluation in Section~\ref{sec:pretrained}, the fine-tuned AgriPINN maintains its strong predictive ability while improving sensitivity to water stress. 

In contrast, the data-driven models show a clear performance drop after fine-tuning. The main reason is that these models are prone to catastrophic forgetting under domain shift. The fine-tuning data, which are specific to water-stress conditions, have different covariate distributions and higher variance than the pretraining data. As a result, the data-driven models overfit to the fine-tuning regime, drifting away from the pretrained feature space. \par

\begin{table}[]
\caption{Performance of AgriPINN and baseline models in predicting winter wheat biomass under three water-stress treatments (sheltered, rainfed, irrigated). Metrics include RMSE (t/ha), coefficient of determination ($R^2$), and Pearson correlation coefficient (CC). LINTUL5 is used as the process-based baseline. Bold values indicate the best performance within each treatment.}
\label{tab:1}
\centering
\resizebox{3.2in}{!}{
\begin{tabular}{llccc}
\hline
Treatment                  & Method          & RMSE (t/ha)   & $R^2$         & CC            \\ \hline
\multirow{5}{*}{Shelter}   & LINTUL5\_Sim    & 1.92          & 0.91          & 0.89          \\
                           & AgriPINN        & \textbf{1.83} & \textbf{0.93} & \textbf{0.94} \\
                           & ConvLSTM-ViT    & 2.06          & 0.89          & 0.79          \\
                           & SLTF            & 4.47          & 0.77          & 0.68          \\
                           & CNN-Transformer & 4.61          & 0.83          & 0.74          \\ \hline
\multirow{5}{*}{Rainfed}   & LINTUL5\_Sim    & 4.40          & 0.90          & 0.87          \\
                           & AgriPINN        & \textbf{2.25} & \textbf{0.92} & \textbf{0.91} \\
                           & ConvLSTM-ViT    & 6.82          & 0.82          & 0.81          \\
                           & SLTF            & 9.39          & 0.72          & 0.66          \\
                           & CNN-Transformer & 9.45          & 0.79          & 0.73          \\ \hline
\multirow{5}{*}{Irrigated} & LINTUL5\_Sim    & 3.06          & 0.90          & 0.86          \\
                           & AgriPINN        & \textbf{2.03} & \textbf{0.93} & \textbf{0.91} \\
                           & ConvLSTM-ViT    & 5.88          & 0.82          & 0.81          \\
                           & SLTF            & 8.85          & 0.79          & 0.74          \\
                           & CNN-Transformer & 8.53          & 0.78          & 0.73          \\ \hline
\end{tabular}
}
\end{table}

\begin{figure}[]   
    \centering  
    \includegraphics[width=6in]{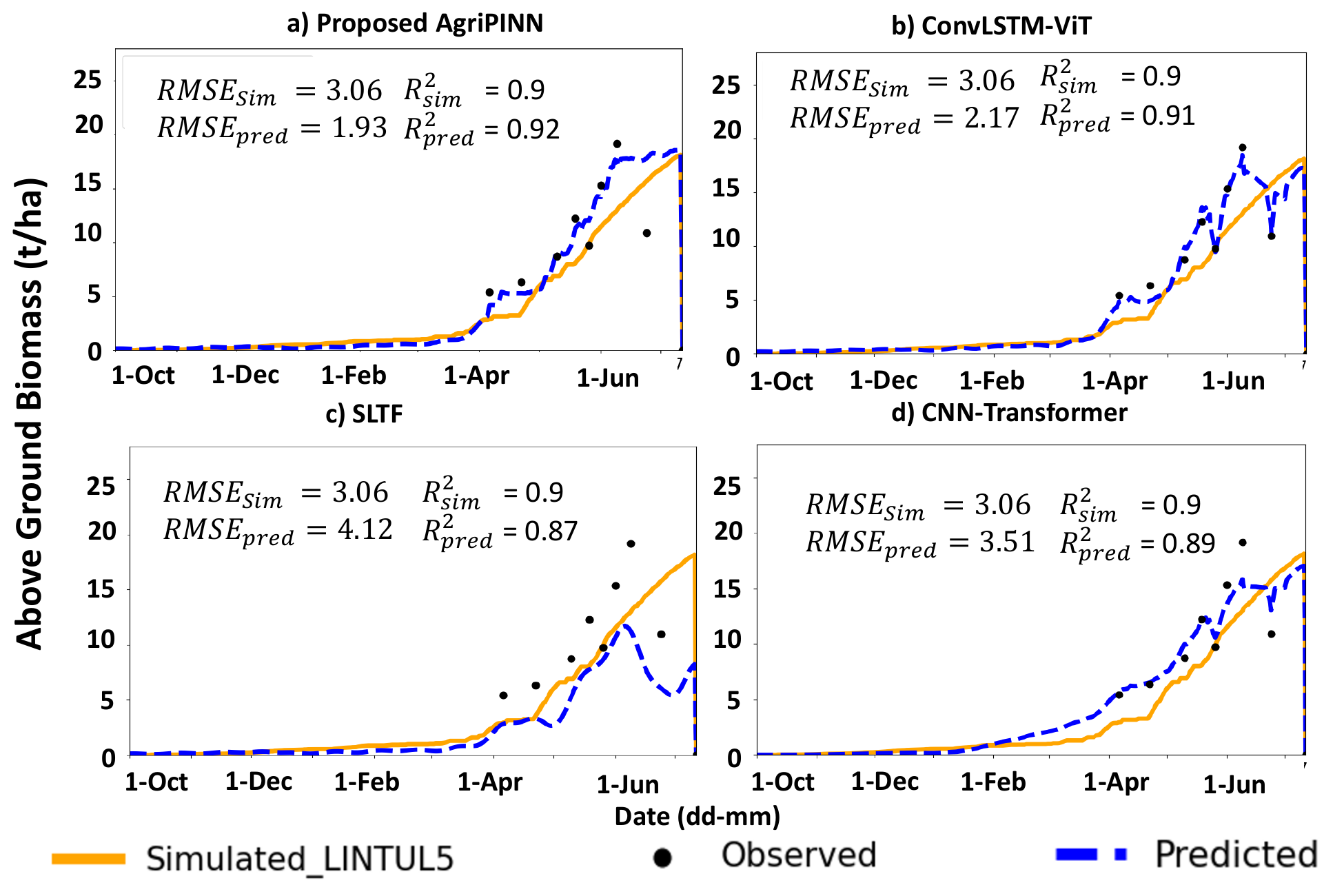}  
    \caption{Temporal trajectories of winter wheat AGB under irrigated water treatment for four modelling approaches (AgriPINN, ConvLSTM–ViT, SLTF, and CNN–Transformer). Black dots indicate in-situ observations, orange lines show LINTUL5 simulations, and blue dashed lines denote model predictions. AgriPINN reproduces the observed AGB dynamics more accurately than the data-driven baselines and captures treatment-specific growth reductions, whereas baseline models fail to differentiate the water-stress scenarios. More detailed comparisons across different water treatments can be found in the Supplementary Materials. }
    \label{fig:4}  
\end{figure}

Figure~\ref{fig:4} illustrates the temporal trajectories of winter wheat AGB under different water treatments for four modelling approaches (AgriPINN, ConvLSTM–ViT, SLTF, and CNN–Transformer). Each panel compares in-situ observations (black dots), LINTUL5 simulations (orange line), and model predictions (blue dashed line). As shown in Fig.~\ref{fig:4}a, AgriPINN closely matches the observed dynamics, accurately capturing the onset of rapid growth, the steep increase during stem elongation, and the timing and magnitude of peak AGB, which is reflected in its lowest RMSE\textsubscript{pred} and highest $R^2_{\text{pred}}$. In contrast, the data-driven baselines (Fig.~\ref{fig:4}b–d) exhibit larger deviations from the observations: ConvLSTM–ViT and SLTF underestimate early-season biomass and misrepresent the curvature around the peak, whereas the CNN–Transformer overestimates AGB during the mid–late growth stages. These discrepancies indicate that, without process constraints, baseline models struggle to reproduce the full temporal pattern of biomass accumulation even under a single irrigated regime. More detailed comparisons across different water treatments can be found in the Supplementary Materials section 7. \par

\begin{figure}[]   
    \centering  
    \includegraphics[width=5.5in]{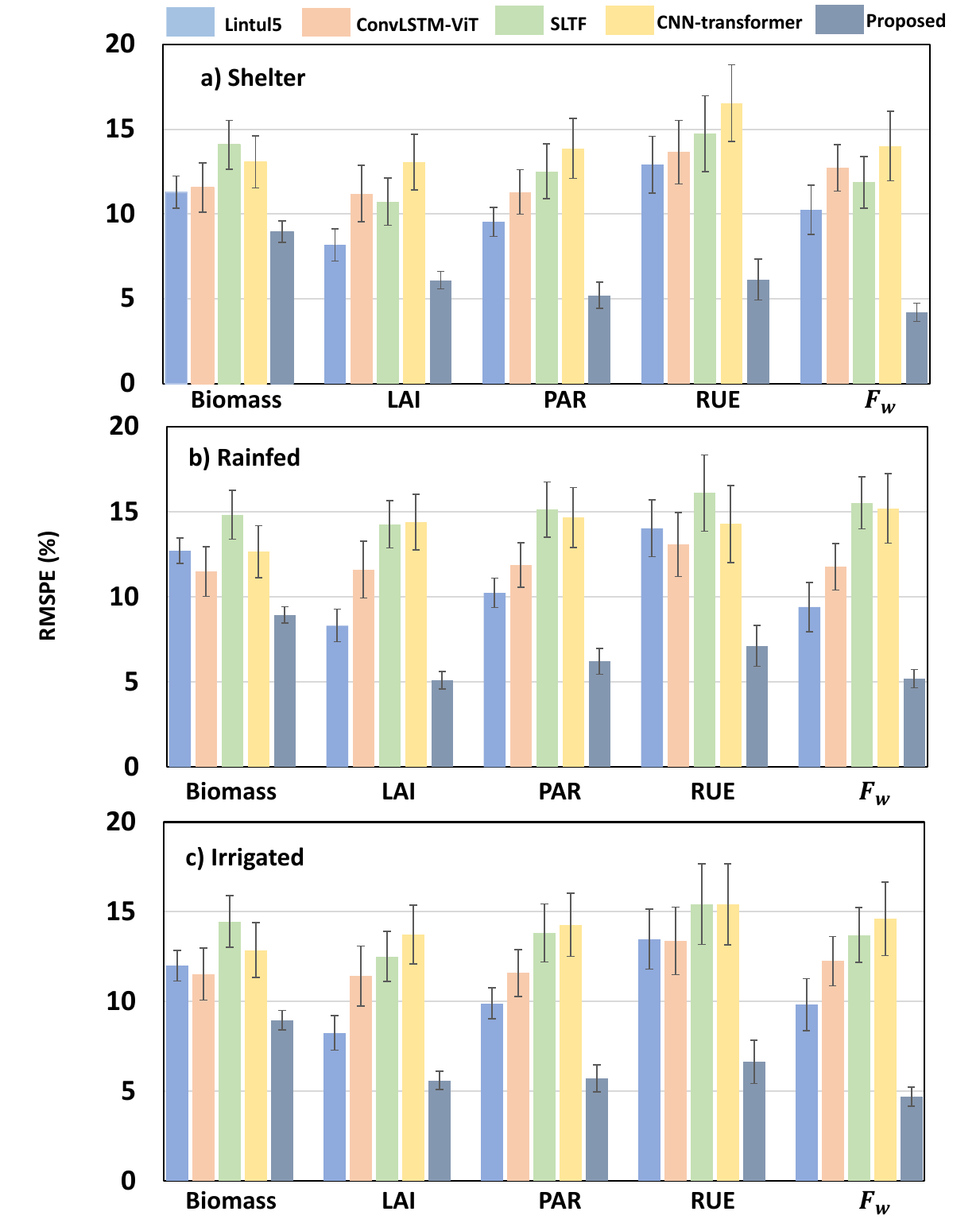}  
    \caption{RMSPE (\%) for AGB and latent physiological variables (LAI, PAR, RUE, $F_W$) under three water-stress treatments (sheltered, rainfed, irrigated). Error bars indicate standard deviations across all experimental plots. AgriPINN achieves consistently lower RMSPE and reduced variability across treatments compared with LINTUL5 and the data-driven baselines, demonstrating improved robustness in both biomass prediction and latent-variable reconstruction.}  
    \label{fig:5}  
\end{figure}

Aside from the overall comparison of the five approaches, a post-hoc analysis using Tukey’s Honest Significant Difference test is applied, with the significance level set at $p < 0.05$. Figure~\ref{fig:5} illustrates the RMSPE of AGB and its physiological variables (LAI, PAR, RUE, and $F_W$) in the winter wheat scenario. As shown in Fig.~\ref{fig:5}, AgriPINN achieves accurate predictions with lower standard deviations and smaller fluctuations across the different water-stress treatments.

\subsection{Spatial variability retrieval of crop AGB}

\begin{figure}[]   
    \centering  
    \includegraphics[width=5.5in]{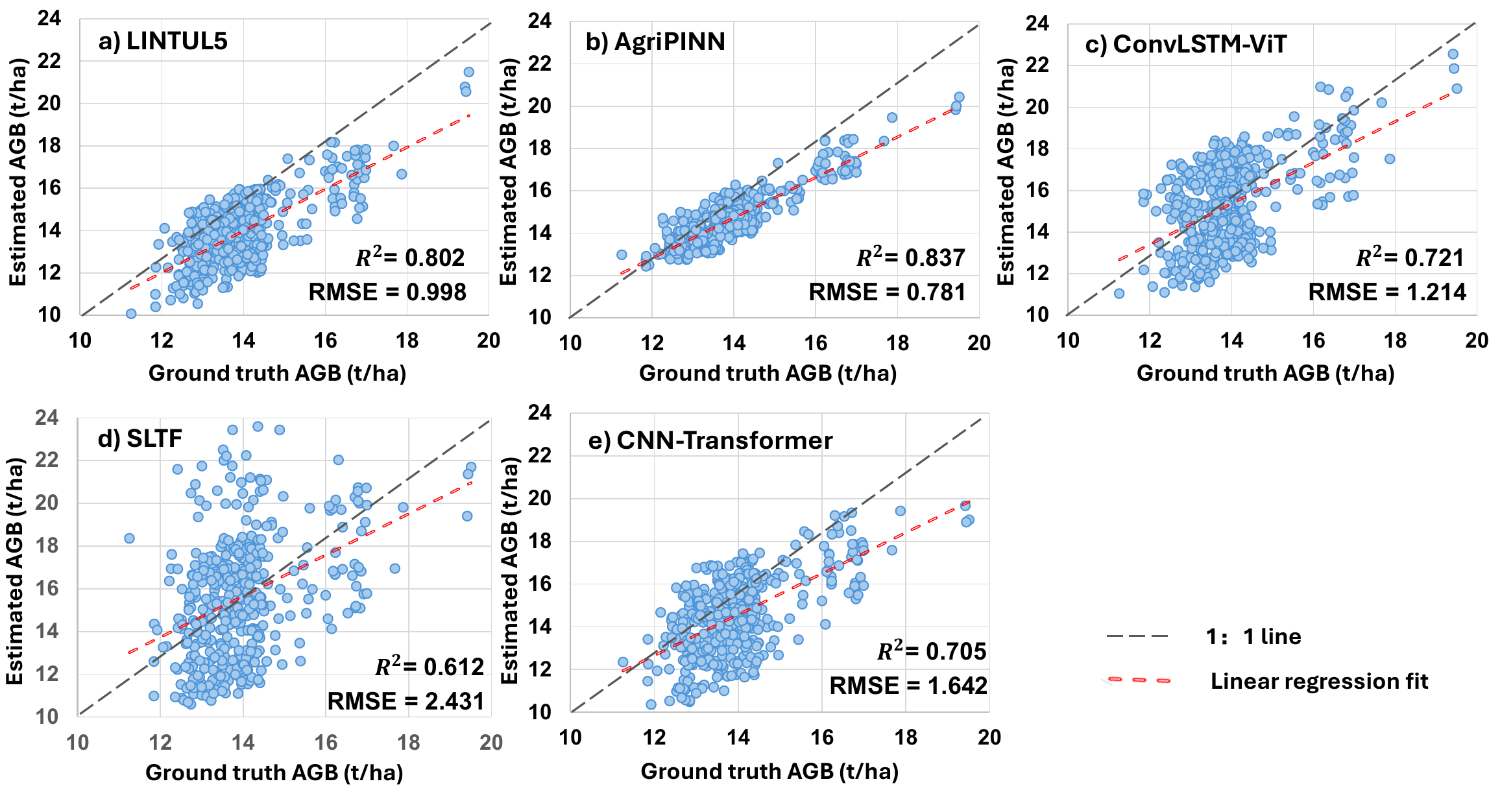}  
    \caption{Regression scatter plots between ground-truth AGB and AGB estimated from (a) LINTUL5, (b) the proposed AgriPINN, (c) ConvLSTM-ViT, (d) SLTF, and (e) CNN-Transformer. Ground-truth AGB is aggregated at the NUTS-3 regional scale (a total of 484 districts), while the estimated AGB values are statistically summarized at the same NUTS-3 level. The proposed model achieves the best predictive performance, with the highest $R^2$ (0.837) and the lowest RMSE (0.781), outperforming both the process-based model (second-best, $R^2 = 0.802$, RMSE $= 0.998$) and the data-driven models ($R^2 < 0.721$, RMSE $> 1.214$).}   
    \label{fig:6_1}  
\end{figure}

To evaluate large-scale spatial retrieval performance, we use ERA5 post-processed daily climate fields (radiation, temperature, precipitation) at 0.25$^\circ$ resolution as model inputs. To increase spatial resolution, the ERA5 data are downscaled to 250\,m using the MODIS MOD09Q1 NDVI product. We use winter wheat in 2016 as a representative example. Figure~\ref{fig:6_1} shows the spatial regression between ground-truth AGB and model-estimated AGB for LINTUL5, AgriPINN, and three data-driven baselines (ConvLSTM–ViT, SLTF, CNN–Transformer). The ground-truth AGB is derived from county-level crop-yield statistics reported in \cite{duden2024german}, converted to biomass using a harvest index of 0.55 as suggested by \cite{rose2017yield}.  \par

Across all models, AgriPINN achieves the best predictive performance, with the highest $R^2$ (0.837) and the lowest RMSE (0.781), outperforming both the process-based LINTUL5 model (second-best, $R^2 = 0.802$, RMSE $= 0.998$) and the purely data-driven baselines ($R^2 < 0.721$, RMSE $> 1.214$). Beyond the summary metrics, the point-cloud patterns in Fig.~\ref{fig:6_1} also reveal important model behaviour. In the LINTUL5 panel (Fig.~\ref{fig:6_1}a), many points cluster along several nearly horizontal bands. This banding pattern indicates that the process-based model responds weakly to spatial variability in climate and management inputs, producing similar AGB estimates for regions with substantially different observed yields. In contrast, AgriPINN (Fig.~\ref{fig:6_1}b) yields a more dispersed cloud of points tightly concentrated around the 1:1 line, demonstrating stronger sensitivity to regional differences while remaining consistent with the biomass-growth law. The data-driven baselines (Fig.~\ref{fig:6_1}c–e) show larger scatter and systematic biases, reflecting either under-sensitivity or overfitting to particular ranges of AGB. Taken together, these results indicate that the proposed AgriPINN model provides the most accurate and spatially responsive AGB predictions when generalized to large-scale applications.

\begin{figure}[]   
    \centering  
    \includegraphics[width=5.5in]{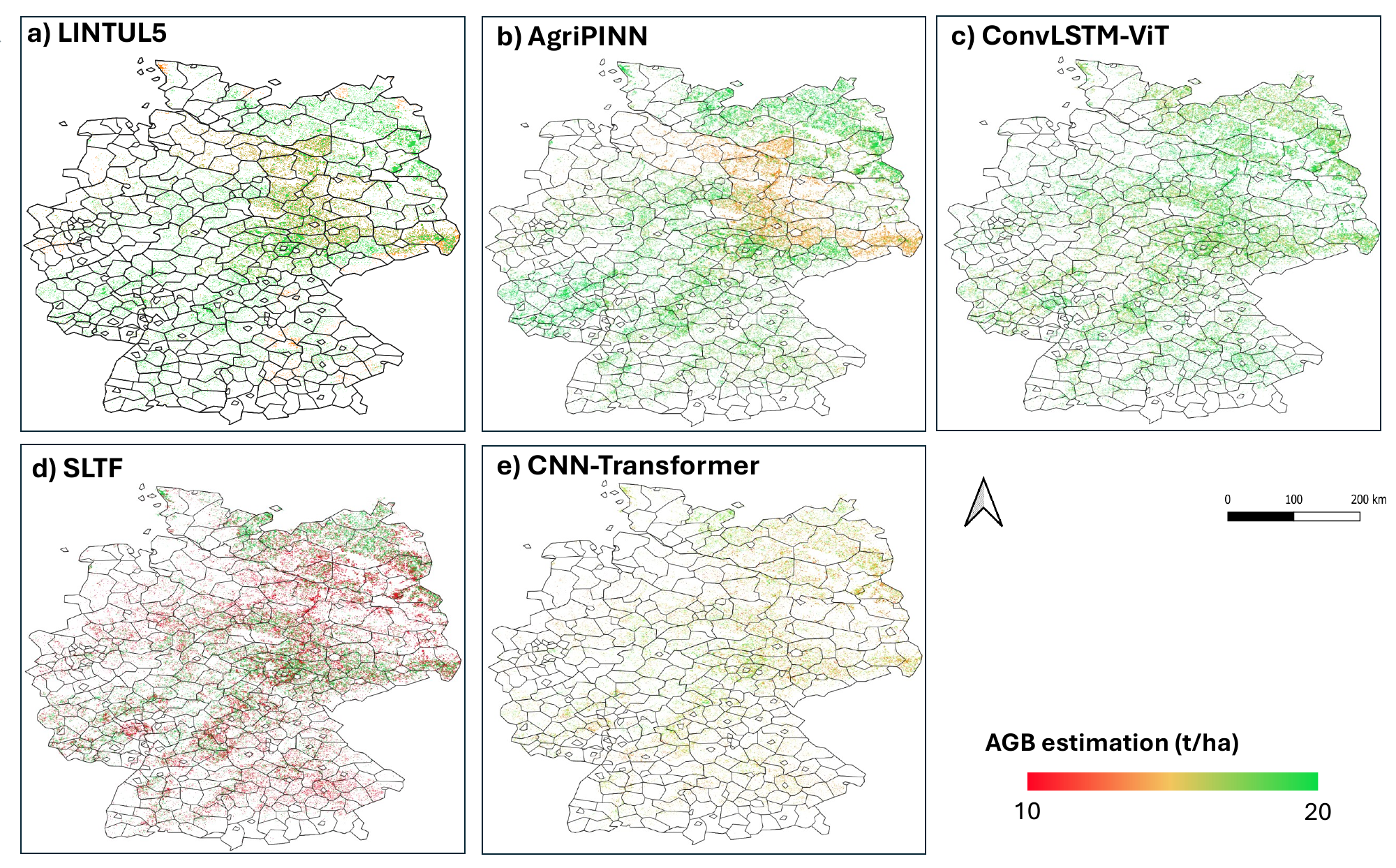}  
    \caption{Spatial distribution of winter-wheat AGB (t/ha) in Germany for 2016 estimated by (a) LINTUL5, (b) the proposed AgriPINN, (c) ConvLSTM–ViT, (d) SLTF, and (e) CNN–Transformer. AGB estimates are produced at 250\,m resolution. Colours indicate AGB magnitude from low (red, $\approx$10\,t/ha) to high (green, $\approx$20\,t/ha). AgriPINN preserves coherent regional gradients comparable to the output of the process-based model LINTUL5 while reducing the overly smooth patterns of the process-based model and the noisy, spatially fragmented structures exhibited by the purely data-driven baselines.} 
    \label{fig:6_2}  
\end{figure}

The spatial variability retrieval of winter wheat AGB across Germany, shown in Fig.~\ref{fig:6_2}, reveals distinct differences among the process-based, hybrid, and data-driven models. The AgriPINN model (Fig.~\ref{fig:6_2}b) demonstrates a spatial distribution of AGB that closely aligns with the process-based LINTUL5 simulation (Fig.~\ref{fig:6_2}a). Both models capture consistent regional patterns, particularly the higher biomass density in the northern and northeastern regions of Germany, as well as the relatively lower biomass values observed in the southern and central regions. This similarity suggests that AgriPINN effectively integrates domain knowledge from process-based modelling, allowing it to retain biologically meaningful spatial structures while enhancing predictive accuracy and highlighting strong yield potentials in the northern lowlands and moderate productivity in central and southern regions. This agreement with empirical evidence further supports the robustness of the hybrid modelling framework, which combines the interpretability of process-based models with the adaptability of deep learning. In contrast, the data-driven models, including ConvLSTM-ViT, SLTF, and CNN-Transformer (Figs.~\ref{fig:6_2}c–e), exhibit weaker consistency in spatial representation and tend to either overestimate or underestimate biomass in several regions, leading to fragmented and less realistic spatial patterns. 

\begin{figure}[]   
    \centering  
    \includegraphics[width=5.5in]{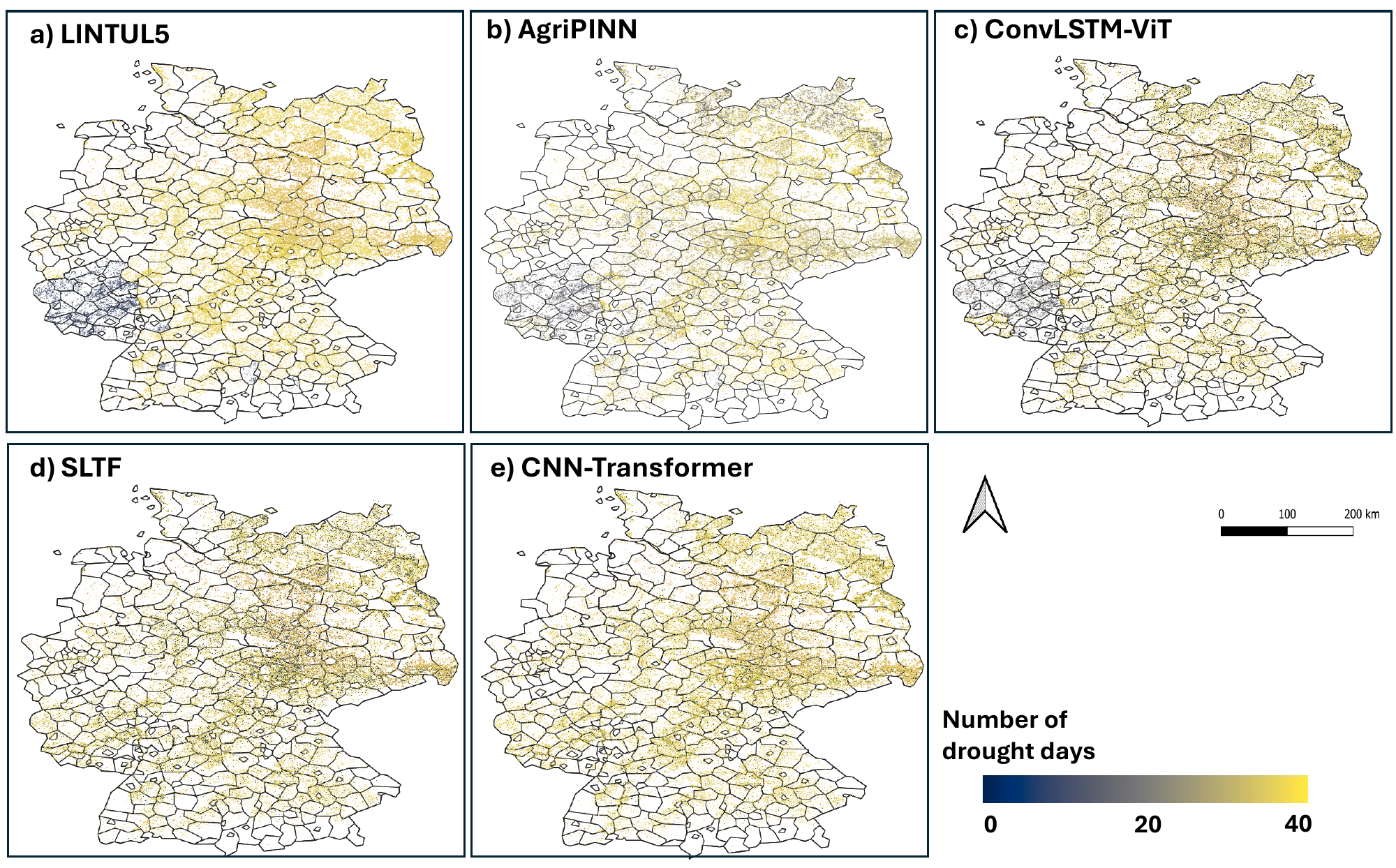}  
    \caption{Spatial distribution of water-stress conditions for winter wheat in Germany during the 2016 growing season, derived from (a) LINTUL5, (b) the proposed AgriPINN, (c) ConvLSTM–ViT, (d) SLTF, and (e) CNN–Transformer. For each model, water stress is summarized as the number of drought days, where higher values (indicated by yellow shades) correspond to more frequent or prolonged periods of water limitation, and lower values (blue shades) indicate few or no drought days. AgriPINN produces spatial patterns that follow large-scale climatic gradients while avoiding the overly strong regional artefacts seen in the process-based LINTUL5 output and the noisier, less structured patterns generated by the purely data-driven baselines.}   
    \label{fig:6_3} 
\end{figure}

The spatial variability of water stress, measured as the number of drought days ($F_W < 0.6$), is presented in Fig.~\ref{fig:6_3}. The proposed AgriPINN (Fig.~\ref{fig:6_3}b) demonstrates a spatial pattern that is highly consistent with the process-based LINTUL5 model (Fig.~\ref{fig:6_3}a). Both models capture the well-documented regional contrasts in drought occurrence across Germany, including the higher frequency of drought days in the northeast and central regions and the comparatively lower water-stress levels in the southwest. This consistency highlights the ability of AgriPINN to effectively incorporate biophysical mechanisms of water balance from process-based modelling while leveraging machine learning flexibility to refine large-scale predictions. By contrast, the data-driven approaches (Figs.~\ref{fig:6_3}c–e) exhibit less coherent spatial representations of water stress. \par

\begin{figure}[]   
    \centering  
    \includegraphics[width=5.5in]{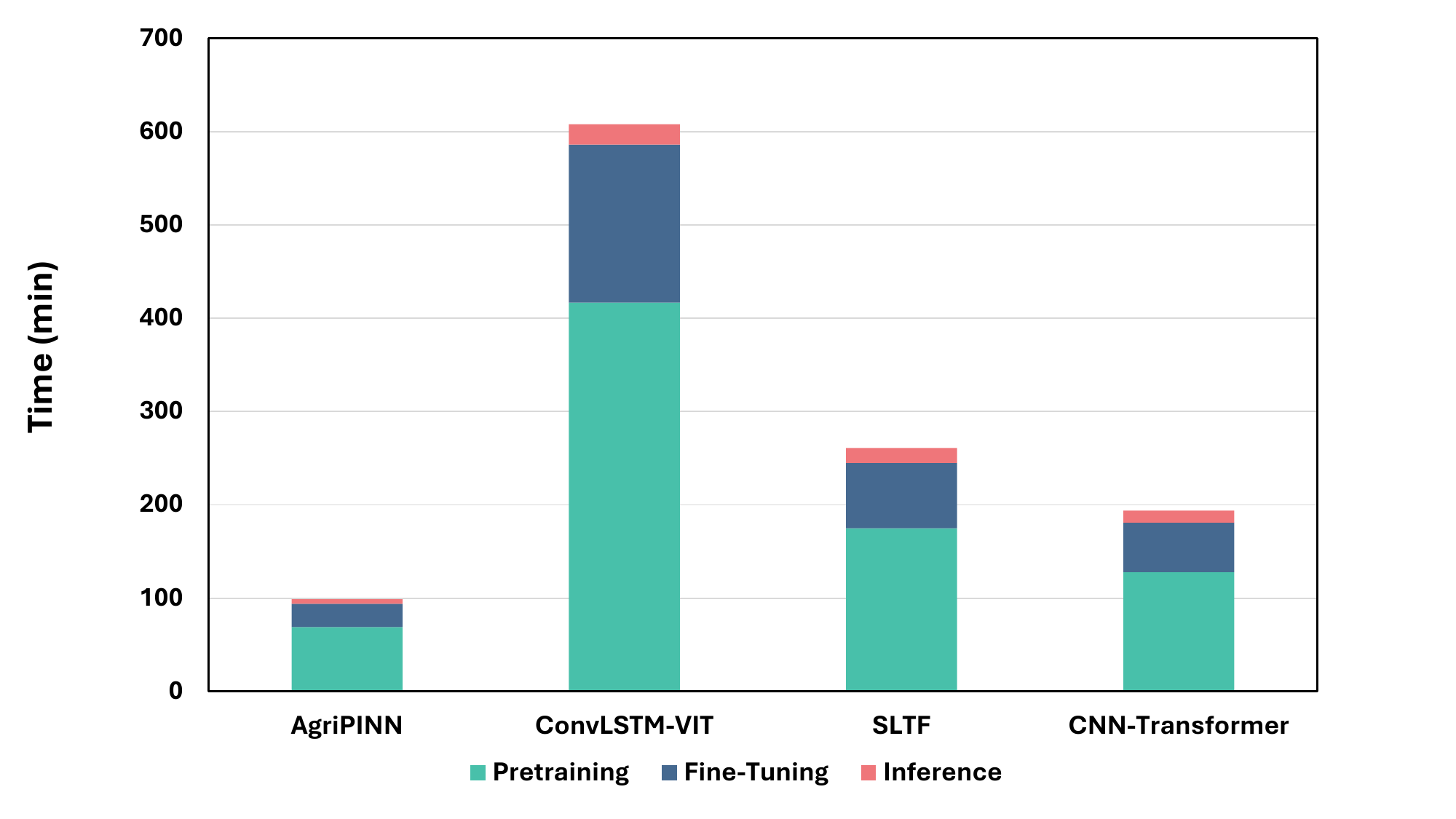}  
    \caption{Comparison of computational efficiency for the deep-learning models. Stacked bars show the total computation time (in min) required for three stages, pretraining (green), fine-tuning (blue), and inference (red), for the proposed AgriPINN, ConvLSTM–ViT, SLTF, and CNN–Transformer. In contrast, AgriPINN achieves the lowest overall computational cost, requiring markedly less pretraining and inference time than both the process-based model and the data-driven baselines.} 
    \label{fig:11_1} 
\end{figure}

The computational efficiency comparison of the spatio-temporal variability, in terms of the stacked time of pretraining, fine-tuning, and inference, is illustrated in Fig.~\ref{fig:11_1}. Our comparison indicates that the proposed AgriPINN achieves the highest end-to-end efficiency among all benchmarks, with minimal runtime. The SOTA deepl learning models, due to their complex multi-layer stacked architectures, exhibit a heavy pretraining burden and substantial fine-tuning cost, yielding a total computational expense roughly four to six times higher than that of AgriPINN. These results suggest that AgriPINN offers a balanced compute allocation, thereby improving computational efficiency while maintaining high sensitivity to water stress.  

\subsection{Ablation analysis}

To assess the generality of the hybrid framework, we integrate the process-informed loss with several representative neural backbones, including a 1D-CNN, an LSTM, ResNet-18, and a Time-Series Transformer (TFT). Table~\ref{tab:9} lists the detailed comparison results. Compared with the original versions of each backbone, our hybrid modelling approach adds no extra parameters or computational burden, yet consistently improves both predictive accuracy and training speed. The gains are most pronounced for lightweight models (1D-CNN and LSTM), where biomass predictions across multiple water-stress scenarios improve markedly. For the deeper ResNet-18, the dominant benefit is faster convergence, indicating that the hybrid mechanism scales favourably with model depth. These results demonstrate the approach’s practical significance, methodological rigour, clear novelty, and potential impact on efficient agricultural analytics.

\begin{table}[t]
\centering
\caption{Evaluation of the proposed AgriPINN-based hybrid modelling across different neural backbones. For each backbone, we report model size (number of parameters in millions), training time to convergence (minutes), and RMSE (t/ha) under three water treatments (shelter, rainfed, irrigated) for the original purely data-driven model (\emph{Original}) and the corresponding process-informed hybrid (\emph{Hybrid}).}
\label{tab:9}
\resizebox{\textwidth}{!}{
\begin{tabular}{lcccccc}
\hline
Backbone & Variant & \# Parameters (M) & Convergence time (min) & RMSE (shelter, t/ha) & RMSE (rainfed, t/ha) & RMSE (irrigated, t/ha) \\
\hline
\multirow{2}{*}{1D-CNN}    
  & Original & 0.7  & 11  & 6.74 & 5.32 & 6.85 \\
  & Hybrid   & 0.7  & 10  & 1.83 & 2.25 & 2.03 \\
\hline
\multirow{2}{*}{LSTM}      
  & Original & 0.5  &  8  & 5.59 & 6.32 & 5.58 \\
  & Hybrid   & 0.5  &  7  & 1.86 & 1.98 & 2.18 \\
\hline
\multirow{2}{*}{ResNet-18} 
  & Original & 11.7 & 492 & 2.17 & 2.03 & 2.56 \\
  & Hybrid   & 11.7 & 376 & 1.81 & 1.89 & 1.92 \\
\hline
\multirow{2}{*}{TFT}       
  & Original & 5.7  & 176 & 2.23 & 2.92 & 2.47 \\
  & Hybrid   & 5.7  & 129 & 1.88 & 1.93 & 1.98 \\
\hline
\end{tabular}
}
\\[0.5ex]
\footnotesize \textit{Note:} ``Original'' denotes the baseline backbone trained without process constraints. ``Hybrid'' denotes the same backbone trained within the AgriPINN framework using the LINTUL5-based process residual as an additional loss term. Convergence time refers to the total wall-clock training time required for each model to reach its best validation performance.
\end{table}

\section{Discussions}
\label{sec:discussion}

In this section, we discuss the performance, generalization capability, and robustness of the proposed AgriPINN framework. We also address its current limitations and outline potential directions for future hybrid agricultural modelling.

\subsection{Performance of AgriPINN in learning process-based biomass dynamics}

The core contribution of AgriPINN lies in its ability to integrate process-based rules into deep learning architectures without incurring the heavy computational costs associated with traditional biophysical simulations. As evidenced by the model performance evaluation in Section~\ref{sec:pretrained}, the proposed framework achieves superior convergence and predictive accuracy compared with state-of-the-art data-driven models. By embedding process-based information as a loss term, AgriPINN effectively regularises the underlying CNN backbone, mitigating the common issue of overfitting to noisy training data.\par

Furthermore, the computational-efficiency analysis (Fig.~\ref{fig:11_1}) highlights a critical advantage. Although a direct comparison is difficult, since process-based models require iterative CPU computations while AgriPINN benefits from GPU acceleration, process-based models frequently become performance bottlenecks in high-resolution, large-scale simulations due to their iterative nature. In contrast, AgriPINN significantly lowers both training time and parameter count. Taken together, these results indicate that the hybrid paradigm retains the biophysical consistency of the process-based ODE while leveraging the inference speed and representational power of deep neural networks.

\subsection{Generalization of AgriPINN}

A key finding from our ablation study (Table~\ref{tab:9}) is the strong generalization capability of the AgriPINN framework across different network backbones. Whether applied to lightweight architectures (1D-CNN, LSTM) or deeper networks (ResNet-18, TFT), the hybrid modelling approach consistently improves performance. Notably, for lightweight models, the hybrid formulation substantially reduces RMSE (e.g., LSTM RMSE under the sheltered treatment decreases from 5.59 to 1.86), indicating that process-based constraints help compensate for limited model capacity. In contrast, for deeper models such as ResNet-18, the primary improvement is faster convergence, with training time reduced from 492\,min to 376\,min.

These results demonstrate that AgriPINN is not merely a specific model architecture but a flexible, generic paradigm. It introduces no additional parameters at inference time, yet enhances the model's ability to learn domain-informed features. This flexibility allows the framework to be adapted to various scenarios; for instance, the backbone can be switched to LSTMs for purely temporal tasks or extended to Transformers for complex sequence modelling, while still enforcing the biophysical constraints defined by the process-based component.

\subsection{Interpretability and Unsupervised Latent Variable Discovery}

A persistent challenge in agricultural deep learning, as noted in recent comprehensive reviews by Jabed \textit{et al}\cite{jabed2024crop}, is the "black box" nature of models like CNNs. While these models excel at pattern matching, they often fail to disentangle causal drivers or provide the physiological transparency required for agronomic decision-making. 
AgriPINN addresses this by successfully recovering latent physiological variables, for this case, LAI, RUE, and $F_w$, without direct supervision. Our results show that the inferred water-stress factor ($F_w$) aligns spatially with drought patterns in Germany (Fig.~\ref{fig:6_3}), providing a transparent explanation for yield suppression in the central and southern regions. The low RMSPE for these latent variables across sheltered, rainfed, and irrigated treatments (Fig.~\ref{fig:5}) confirms that the model is learning valid agronomic mechanisms rather than exploiting statistical artifacts.

\subsection{Robustness to Distribution Shift and Data Assimilation}

Purely data-driven models are prone to catastrophic forgetting and poor out-of-distribution (OOD) generalization. Our experiments revealed that SOTA baselines (ConvLSTM-ViT, SLTF, CNN-Transformer) suffered significant performance degradation when fine-tuned on specific water-stress treatments (e.g., RMSE increasing from ~2.0 t/ha to >4.0 t/ha for SLTF in sheltered conditions). AgriPINN exhibited superior stability under these domain shifts. By constraining the hypothesis space to physically valid trajectories, the process-informed loss acts as a "biophysical constraint." This contrasts with data assimilation approaches, such as Ren \textit{et al.} \cite{ren2024based} or Tewes \textit{et al.} \cite{tewes2020assimilation}, which adjust state variables within a process model using external observations. While effective, data assimilation typically involves computationally intensive optimization loops at every time step. AgriPINN achieves similar biophysical consistency through its differentiable architecture, allowing for rapid, end-to-end inference that is robust to data scarcity and changing environmental regimes.

\subsection{Robustness of spatio-temporal dynamics simulation of crop AGB under water stress}

The robustness of AgriPINN is most evident in its ability to simulate spatio-temporal dynamics under varying water-stress conditions, a setting in which purely data-driven models frequently fail due to domain shift.

\begin{itemize}
\item \textbf{Temporal precision:} As shown in Fig.~\ref{fig:4}, data-driven models suffer from catastrophic forgetting when fine-tuned on specific water treatments, leading to overfitting and poor generalization. In contrast, AgriPINN maintains high sensitivity to water stress, accurately capturing phenological shifts in AGB accumulation—specifically the distinct peaks during the heading (mid-April) and flowering (early June) stages.
\item \textbf{Spatial heterogeneity:} At the national scale, the spatial retrieval of winter wheat AGB across Germany (Fig.~\ref{fig:6_2}) confirms that AgriPINN captures regional heterogeneity consistent with both ground truth and process-based simulations. It correctly identifies high-yield potential in the northern lowlands (e.g., Schleswig-Holstein) and drought-induced suppression in eastern and central regions. This spatial agreement ($R^2 = 0.837$) outperforms the data-driven baselines, which produce more fragmented and less realistic spatial patterns.
\end{itemize}

Overall, these findings indicate that the hybrid loss guides the model toward biophysically consistent AGB evolution, supporting robust predictions even when the data distribution shifts between the general pretraining phase and the water-stress-specific fine-tuning stage.

\subsection{Limitations and future work}

Despite the promising results, several limitations of the current framework warrant further investigation. First, the accuracy of the physical constraint depends on the governing equations of the LINTUL5 model; if the underlying process descriptions are oversimplified or biased, the hybrid model may inherit these limitations. Second, while the model performs well for biomass and water stress, it has not yet been evaluated for other biotic or abiotic stressors such as pest outbreaks or nutrient deficiencies. 

Future research will focus on three main directions:
\begin{enumerate}
\item \textbf{Expansion of biophysical constraints:} We aim to incorporate more complex agronomic rules, such as detailed nutrient cycling (N–P–K balances) and soil-conservation principles, to enhance the simulation of multi-crop rotation systems.
\item \textbf{Multi-objective optimization:} We plan to extend the hybrid loss formulation to support multiple agronomic targets simultaneously, enabling joint prediction of biomass, yield, irrigation demand, soil moisture, and potentially carbon-sequestration indicators.
\item \textbf{Real-world deployment:} To bridge the gap between research and practice, we intend to optimize AgriPINN for real-time inference on edge devices and develop user-friendly interfaces. This will allow farmers and agricultural managers to utilise the proposed high-efficiency framework for informed decision-making in precision agriculture.
\end{enumerate}

\section{Conclusion}
\label{sec:conclusion}
In this work, we presented AgriPINN, a generic hybrid framework for estimating the spatio-temporal variability of crop Above-Ground Biomass (AGB). By bridging the gap between mechanistic modeling and data-driven learning, AgriPINN addresses the limitations of both paradigms: it overcomes the computational bottlenecks of process-based models and the data-hungriness and overfitting risks of pure deep learning models. Methodologically, we enforced a mechanistic growth through a differentiable residual term, enabling the model to generate physiological variables (LAI, PAR, RUE, $F_W$) in an unsupervised manner to cross-validate AGB predictions.

Extensive experiments demonstrate that AgriPINN significantly outperforms SOTA data-driven baselines. Quantitatively, it achieved the highest accuracy across diverse water-stress treatments (average RMSE $\approx$ 2.01, $R^2$ $\approx$ 0.93) and the most accurate large-scale spatial retrieval ($R^2 = 0.837$). Phenomenologically, the model successfully captured critical agronomic dynamics. Temporally, it identified that water stress exerts the strongest impact on yield formation between mid-April and early June (heading to flowering stages). Spatially, AgriPINN correctly characterized regional heterogeneity, reflecting higher biomass density in the well-irrigated North/Northeast of Germany and drought-induced suppression in the Central/Southern regions—patterns that pure data-driven models failed to reconstruct coherently.

Furthermore, our ablation study confirms that AgriPINN is an architecture-agnostic and modular framework. Whether applied to lightweight CNNs or deep ResNet architectures, the hybrid approach consistently improved convergence rates and predictive accuracy. Crucially, AgriPINN offers a scalable solution for real-world applications, reducing computational resource consumption by a factor of eight compared to traditional process-based simulations. This efficiency, combined with its robustness against catastrophic forgetting, positions AgriPINN as a viable engine for real-time agricultural digital twins.

Future work will explore the auto-tuning of hyperparameters (e.g., loss weight $\lambda$) to reduce manual intervention and incorporate more complex biophysical processes, such as nutrient cycling, to broaden the model's applicability. Additionally, integrating uncertainty quantification will be a priority to provide confidence intervals essential for risk-aware decision-making. Overall, AgriPINN demonstrates that by respecting the laws of nature within a learning framework, we can achieve predictions that are not only accurate and efficient but also interpretable and biologically grounded, empowering the next generation of sustainable agricultural management.

\section*{Acknowledgments}
This work was supported in part by the Biotechnology and Biological Sciences Research Council (BBSRC) under Grant BB/Y513763/1, and in part by Engineering and Physical Sciences Research Council (EPSRC) under Grant EP/X013707/1, and the Leibniz Association under LL-SYSTAIN (Grant Labs-2024-IÖR), and the German National Science Foundation (DFG) under FAIRagro (Grant NFDI 51/1). (Corresponding author: Liangxiu Han).

\bibliographystyle{unsrt}  
\bibliography{references}

@article{van2020crop,
  title={Crop yield prediction using machine learning: A systematic literature review},
  author={Van Klompenburg, Thomas and Kassahun, Ayalew and Catal, Cagatay},
  journal={Computers and electronics in agriculture},
  volume={177},
  pages={105709},
  year={2020},
  publisher={Elsevier}
}

@article{dehghanisanij2022hybrid,
  title={A hybrid machine learning approach for estimating the water-use efficiency and yield in agriculture},
  author={Dehghanisanij, Hossein and Emami, Hojjat and Emami, Somayeh and Rezaverdinejad, Vahid},
  journal={Scientific Reports},
  volume={12},
  number={1},
  pages={6728},
  year={2022},
  publisher={Nature Publishing Group UK London}
}

@article{paudel2021machine,
  title={Machine learning for large-scale crop yield forecasting},
  author={Paudel, Dilli and Boogaard, Hendrik and de Wit, Allard and Janssen, Sander and Osinga, Sjoukje and Pylianidis, Christos and Athanasiadis, Ioannis N},
  journal={Agricultural Systems},
  volume={187},
  pages={103016},
  year={2021},
  publisher={Elsevier}
}

@article{cheng2024multimodal,
  title={A Multimodal and Temporal Network-Based Yield Assessment Method for Different Heat-Tolerant Genotypes of Wheat},
  author={Cheng, Tianyu and Li, Min and Quan, Longzhe and Song, Youhong and Lou, Zhaoxia and Li, Hailong and Du, Xiaocao},
  journal={Agronomy},
  volume={14},
  number={8},
  pages={1694},
  year={2024},
  publisher={MDPI}
}

@article{ren2024based,
  title={Based on historical weather data to predict summer field-scale maize yield: Assimilation of remote sensing data to WOFOST model by ensemble Kalman filter algorithm},
  author={Ren, Shixiong and Chen, Hao and Hou, Jian and Zhao, Peng and Feng, Hao and others},
  journal={Computers and Electronics in Agriculture},
  volume={219},
  pages={108822},
  year={2024},
  publisher={Elsevier}
}

@article{duden2024german,
  title={German yield and area data for 11 crops from 1979 to 2021 at a harmonized spatial resolution of 397 districts},
  author={Duden, Christoph and Nacke, Christina and Offermann, Frank},
  journal={Scientific Data},
  volume={11},
  number={1},
  pages={95},
  year={2024},
  publisher={Nature Publishing Group UK London}
}

@article{nguyen2024multi,
  title={Multi-year aboveground data of minirhizotron facilities in Selhausen},
  author={Nguyen, Thuy Huu and Lopez, Gina and Seidel, Sabine J and L{\"a}rm, Lena and Bauer, Felix Maximilian and Klotzsche, Anja and Schnepf, Andrea and Gaiser, Thomas and H{\"u}ging, Hubert and Ewert, Frank},
  journal={Scientific Data},
  volume={11},
  number={1},
  pages={674},
  year={2024},
  publisher={Nature Publishing Group UK London}
}

@article{mirhoseininejad2024convlstm,
  title={ConvLSTM-ViT: A Deep Neural Network for Crop Yield Prediction Using Earth Observations and Remotely Sensed Data},
  author={MirhoseiniNejad, S Mahdi and Abbasi-Moghadam, Darioush and Sharifi, Aireza},
  journal={IEEE Journal of Selected Topics in Applied Earth Observations and Remote Sensing},
  year={2024},
  publisher={IEEE}
}

@article{du2025enhancing,
  title={Enhancing Winter Wheat Yield Estimation With a CNN-Transformer Hybrid Framework Utilizing Multiple Remotely Sensed Parameters},
  author={Du, Jiangli and Zhang, Yue and Wang, Pengxin and Tansey, Kevin and Liu, Junming and Zhang, Shuyu},
  journal={IEEE Transactions on Geoscience and Remote Sensing},
  year={2025},
  publisher={IEEE}
}

@article{guo2024novel,
  title={A novel transformer-based neural network under model interpretability for improving wheat yield estimation using remotely sensed multi-variables},
  author={Guo, Fengwei and Wang, Pengxin and Tansey, Kevin and Zhang, Yue and Li, Mingqi and Liu, Junming and Zhang, Shuyu},
  journal={Computers and Electronics in Agriculture},
  volume={223},
  pages={109111},
  year={2024},
  publisher={Elsevier}
}

@article{wolf2012user,
  title={User guide for LINTUL5: Simple generic model for simulation of crop growth under potential, water limited and nitrogen, phosphorus and potassium limited conditions.},
  author={Wolf, J},
  journal={Wageningen University},
  year={2012},
  publisher={Wageningen University}
}

@article{shahhosseini2021coupling,
  title={Coupling machine learning and crop modeling improves crop yield prediction in the US Corn Belt},
  author={Shahhosseini, Mohsen and Hu, Guiping and Huber, Isaiah and Archontoulis, Sotirios V},
  journal={Scientific reports},
  volume={11},
  number={1},
  pages={1606},
  year={2021},
  publisher={Nature Publishing Group UK London}
}

@article{tewes2020methods,
  title={How do methods assimilating Sentinel-2-derived LAI combined with two different sources of soil input data affect the crop model-based estimation of wheat biomass at sub-field level?},
  author={Tewes, Andreas and Hoffmann, Holger and Nolte, Manuel and Krauss, Gunther and Sch{\"a}fer, Fabian and Kerkhoff, Christian and Gaiser, Thomas},
  journal={Remote sensing},
  volume={12},
  number={6},
  pages={925},
  year={2020},
  publisher={MDPI}
}

@article{liu2024knowledge,
  title={Knowledge-guided machine learning can improve carbon cycle quantification in agroecosystems},
  author={Liu, Licheng and Zhou, Wang and Guan, Kaiyu and Peng, Bin and Xu, Shaoming and Tang, Jinyun and Zhu, Qing and Till, Jessica and Jia, Xiaowei and Jiang, Chongya and others},
  journal={Nature communications},
  volume={15},
  number={1},
  pages={357},
  year={2024},
  publisher={Nature Publishing Group UK London}
}

@article{rose2017yield,
  title={Yield formation of Central-European winter wheat cultivars on a large scale perspective},
  author={Rose, T and Nagler, S and Kage, H},
  journal={European Journal of Agronomy},
  volume={86},
  pages={93--102},
  year={2017},
  publisher={Elsevier}
}

@article{worrall2021domain,
  title={Domain-guided machine learning for remotely sensed in-season crop growth estimation},
  author={Worrall, George and Rangarajan, Anand and Judge, Jasmeet},
  journal={Remote Sensing},
  volume={13},
  number={22},
  pages={4605},
  year={2021},
  publisher={MDPI}
}

@article{kellner2019response,
  title={Response of maize biomass and soil water fluxes on elevated CO2 and drought—from field experiments to process-based simulations},
  author={Kellner, Juliane and Houska, Tobias and Manderscheid, Remy and Weigel, Hans-Joachim and Breuer, Lutz and Kraft, Philipp},
  journal={Global change biology},
  volume={25},
  number={9},
  pages={2947--2957},
  year={2019},
  publisher={Wiley Online Library}
}

@article{ding2015modeling,
  title={Modeling crop water use in an irrigated maize cropland using a biophysical process-based model},
  author={Ding, Risheng and Kang, Shaozhong and Du, Taisheng and Hao, Xinmei and Tong, Ling},
  journal={Journal of Hydrology},
  volume={529},
  pages={276--286},
  year={2015},
  publisher={Elsevier}
}

@article{jones2003dssat,
  title={The DSSAT cropping system model},
  author={Jones, James W and Hoogenboom, Gerrit and Porter, Cheryl H and Boote, Ken J and Batchelor, William D and Hunt, L Allen and Wilkens, Paul W and Singh, Upendra and Gijsman, Arjan J and Ritchie, Joe T},
  journal={European journal of agronomy},
  volume={18},
  number={3-4},
  pages={235--265},
  year={2003},
  publisher={Elsevier}
}

@article{keating2003overview,
  title={An overview of APSIM, a model designed for farming systems simulation},
  author={Keating, Brian A and Carberry, Peter S and Hammer, Graeme L and Probert, Mervyn E and Robertson, Michael J and Holzworth, Dean and Huth, Neil I and Hargreaves, John NG and Meinke, Holger and Hochman, Zvi and others},
  journal={European journal of agronomy},
  volume={18},
  number={3-4},
  pages={267--288},
  year={2003},
  publisher={Elsevier}
}

@article{gumma2020agricultural,
  title={Agricultural cropland extent and areas of South Asia derived using Landsat satellite 30-m time-series big-data using random forest machine learning algorithms on the Google Earth Engine cloud},
  author={Gumma, Murali Krishna and Thenkabail, Prasad S and Teluguntla, Pardhasaradhi G and Oliphant, Adam and Xiong, Jun and Giri, Chandra and Pyla, Vineetha and Dixit, Sreenath and Whitbread, Anthony M},
  journal={GIScience \& Remote Sensing},
  volume={57},
  number={3},
  pages={302--322},
  year={2020},
  publisher={Taylor Francis}
}

@inproceedings{nivetha2024optimizing,
  title={Optimizing crop yields with multilayer perceptrons: A data-driven approach},
  author={Nivetha, N and Usharani, S},
  booktitle={2024 8th International Conference on Inventive Systems and Control (ICISC)},
  pages={295--300},
  year={2024},
  organization={IEEE}
}

@article{el2025review,
  title={A review of CNN applications in smart agriculture using multimodal data},
  author={El Sakka, Mohammad and Ivanovici, Mihai and Chaari, Lotfi and Mothe, Josiane},
  journal={Sensors},
  volume={25},
  number={2},
  pages={472},
  year={2025},
  publisher={MDPI}
}

@article{xie2024recent,
  title={Recent advances in Transformer technology for agriculture: A comprehensive survey},
  author={Xie, Weijun and Zhao, Maocheng and Liu, Ying and Yang, Deyong and Huang, Kai and Fan, Chenlong and Wang, Zhandong},
  journal={Engineering Applications of Artificial Intelligence},
  volume={138},
  pages={109412},
  year={2024},
  publisher={Elsevier}
}

@article{hu2023ai,
  title={An AI framework integrating physics-informed neural network with predictive control for energy-efficient food production in the built environment},
  author={Hu, Guoqing and You, Fengqi},
  journal={Applied Energy},
  volume={348},
  pages={121450},
  year={2023},
  publisher={Elsevier}
}

@article{dhillon2023integrating,
  title={Integrating random forest and crop modeling improves the crop yield prediction of winter wheat and oil seed rape},
  author={Dhillon, Maninder Singh and Dahms, Thorsten and Kuebert-Flock, Carina and Rummler, Thomas and Arnault, Joel and Steffan-Dewenter, Ingolf and Ullmann, Tobias},
  journal={Frontiers in Remote Sensing},
  volume={3},
  pages={1010978},
  year={2023},
  publisher={Frontiers Media SA}
}

@article{nakajima2023biomass,
  title={Biomass estimation of World rice (Oryza sativa L.) core collection based on the convolutional neural network and digital images of canopy},
  author={Nakajima, Kota and Tanaka, Yu and Katsura, Keisuke and Yamaguchi, Tomoaki and Watanabe, Tomoya and Shiraiwa, Tatsuhiko},
  journal={Plant Production Science},
  volume={26},
  number={2},
  pages={187--196},
  year={2023},
  publisher={Taylor Francis}
}

@article{gafurov2023advancing,
  title={Advancing agricultural crop recognition: the application of LSTM networks and spatial generalization in satellite data analysis},
  author={Gafurov, Artur and Mukharamova, Svetlana and Saveliev, Anatoly and Yermolaev, Oleg},
  journal={Agriculture},
  volume={13},
  number={9},
  pages={1672},
  year={2023},
  publisher={MDPI}
}

@article{jacome2025agritransformer,
  title={AgriTransformer: A Transformer-Based Model with Attention Mechanisms for Enhanced Multimodal Crop Yield Prediction},
  author={Jacome Galarza, Luis and Realpe, Miguel and Vinn-Ludena, Marlon Santiago and Calderon, Maria Fernanda and Jaramillo, Silvia},
  journal={Electronics},
  volume={14},
  number={12},
  pages={2466},
  year={2025},
  publisher={MDPI}
}

@article{seidel2022simulating,
  title={Simulating root growth as a function of soil strength and yield with a field-scale crop model coupled with a 3D architectural root model},
  author={Seidel, Sabine Julia and Gaiser, Thomas and Srivastava, Amit Kumar and Leitner, Daniel and Schmittmann, Oliver and Athmann, Miriam and Kautz, Timo and Guigue, Julien and Ewert, Frank and Schnepf, Andrea},
  journal={Frontiers in Plant Science},
  volume={13},
  pages={865188},
  year={2022},
  publisher={Frontiers Media SA}
}

@article{laniak2013integrated,
  title={Integrated environmental modeling: a vision and roadmap for the future},
  author={Laniak, Gerard F and Olchin, Gabriel and Goodall, Jonathan and Voinov, Alexey and Hill, Mary and Glynn, Pierre and Whelan, Gene and Geller, Gary and Quinn, Nigel and Blind, Michiel and others},
  journal={Environmental modelling \& software},
  volume={39},
  pages={3--23},
  year={2013},
  publisher={Elsevier}
}

@article{nakayama2022impact,
  title={Impact of anthropogenic disturbances on carbon cycle changes in terrestrial-aquatic-estuarine continuum by using an advanced process-based model},
  author={Nakayama, Tadanobu},
  journal={Hydrological Processes},
  volume={36},
  number={2},
  pages={e14471},
  year={2022},
  publisher={Wiley Online Library}
}

@article{macpherson2020linking,
  title={Linking ecosystem services and the SDGs to farm-level assessment tools and models},
  author={MacPherson, Joseph and Paul, Carsten and Helming, Katharina},
  journal={Sustainability},
  volume={12},
  number={16},
  pages={6617},
  year={2020},
  publisher={MDPI}
}

@article{couedel2024long,
  title={Long-term soil organic carbon and crop yield feedbacks differ between 16 soil-crop models in sub-Saharan Africa},
  author={Cou{\"e}del, Antoine and Falconnier, Gatien N and Adam, Myriam and Cardinael, R{\'e}mi and Boote, Kenneth and Justes, Eric and Smith, Ward N and Whitbread, Anthony M and Affholder, Fran{\c{c}}ois and Balkovic, Juraj and others},
  journal={European Journal of Agronomy},
  volume={155},
  pages={127109},
  year={2024},
  publisher={Elsevier}
}

@article{enders2023simplace,
  title={SIMPLACE—a versatile modelling and simulation framework for sustainable crops and agroecosystems},
  author={Enders, Andreas and Vianna, Murilo and Gaiser, Thomas and Krauss, Gunther and Webber, Heidi and Srivastava, Amit Kumar and Seidel, Sabine Julia and Tewes, Andreas and Rezaei, Ehsan Eyshi and Ewert, Frank},
  journal={in silico Plants},
  volume={5},
  number={1},
  pages={diad006},
  year={2023},
  publisher={Oxford University Press UK}
}

@article{mccown1995apsim,
  title={APSIM: an agricultural production system simulation model for operational research},
  author={McCown, RL and Hammer, Graeme L and Hargreaves, John NG and Holzworth, D and Huth, Neil I},
  journal={Mathematics and computers in simulation},
  volume={39},
  number={3-4},
  pages={225--231},
  year={1995},
  publisher={North-Holland}
}

@article{bai2024evaluation,
  title={Evaluation of wheat yield in North China Plain under extreme climate by coupling crop model with machine learning},
  author={Bai, Huizi and Xiao, Dengpan and Tang, Jianzhao and Li Liu, De},
  journal={Computers and Electronics in Agriculture},
  volume={217},
  pages={108651},
  year={2024},
  publisher={Elsevier}
}

@article{jabed2024crop,
  title={Crop yield prediction in agriculture: A comprehensive review of machine learning and deep learning approaches, with insights for future research and sustainability},
  author={Jabed, Md Abu and Murad, Masrah Azrifah Azmi},
  journal={Heliyon},
  volume={10},
  number={24},
  year={2024},
  publisher={Elsevier}
}

@article{tewes2020assimilation,
  title={Assimilation of sentinel-2 estimated LAI into a crop model: influence of timing and frequency of acquisitions on simulation of water stress and biomass production of winter wheat},
  author={Tewes, Andreas and Montzka, Carsten and Nolte, Manuel and Krauss, Gunther and Hoffmann, Holger and Gaiser, Thomas},
  journal={Agronomy},
  volume={10},
  number={11},
  pages={1813},
  year={2020},
  publisher={MDPI}
}

@article{halder2025robust,
  title={A robust and scalable crop mapping framework using advanced machine learning and optical and SAR imageries},
  author={Halder, Krishnagopal and Srivastava, Amit Kumar and Zheng, Wenzhi and Alsafadi, Karam and Zhao, Gang and Maerker, Michael and Singh, Manmeet and Guoging, Lei and Ghosh, Anitabha and Vianna, Murilo and others},
  journal={Smart Agricultural Technology},
  pages={101354},
  year={2025},
  publisher={Elsevier}
}

@article{nguyen2022responses,
  title={Responses of winter wheat and maize to varying soil moisture: From leaf to canopy},
  author={Nguyen, Thuy Huu and Langensiepen, Matthias and Gaiser, Thomas and Webber, Heidi and Ahrends, Hella and Hueging, Hubert and Ewert, Frank},
  journal={Agricultural and Forest Meteorology},
  volume={314},
  pages={108803},
  year={2022},
  publisher={Elsevier}
}

@article{kheir2023integrating,
  title={Integrating APSIM model with machine learning to predict wheat yield spatial distribution},
  author={Kheir, Ahmed MS and Mkuhlani, Siyabusa and Mugo, Jane W and Elnashar, Abdelrazek and Nangia, Vinay and Devare, Medha and Govind, Ajit},
  journal={Agronomy Journal},
  volume={115},
  number={6},
  pages={3188--3196},
  year={2023},
  publisher={Wiley Online Library}
}

@article{droutsas2022integration,
  title={Integration of machine learning into process-based modelling to improve simulation of complex crop responses},
  author={Droutsas, Ioannis and Challinor, Andrew J and Deva, Chetan R and Wang, Enli},
  journal={in silico Plants},
  volume={4},
  number={2},
  pages={diac017},
  year={2022},
  publisher={Oxford University Press UK}
}

@article{van2003approaches,
  title={On approaches and applications of the Wageningen crop models},
  author={van Ittersum, Martin K and Leffelaar, Peter A and Van Keulen, H and Kropff, Martin J and Bastiaans, Lammert and Goudriaan, Jan},
  journal={European journal of agronomy},
  volume={18},
  number={3-4},
  pages={201--234},
  year={2003},
  publisher={Elsevier}
}

@article{van1989wofost,
  title={WOFOST: a simulation model of crop production},
  author={Van Diepen, CA van and Wolf, J van and Van Keulen, H and Rappoldt, C},
  journal={Soil use and management},
  volume={5},
  number={1},
  pages={16--24},
  year={1989},
  publisher={Wiley Online Library}
}

@article{holzworth2014apsim,
  title={APSIM--evolution towards a new generation of agricultural systems simulation},
  author={Holzworth, Dean P and Huth, Neil I and deVoil, Peter G and Zurcher, Eric J and Herrmann, Neville I and McLean, Greg and Chenu, Karine and van Oosterom, Erik J and Snow, Val and Murphy, Chris and others},
  journal={Environmental Modelling \& Software},
  volume={62},
  pages={327--350},
  year={2014},
  publisher={Elsevier}
}

@article{asseng2013uncertainty,
  title={Uncertainty in simulating wheat yields under climate change},
  author={Asseng, Senthold and Ewert, Frank and Rosenzweig, Cynthia and Jones, James W and Hatfield, Jerry L and Ruane, Alex C and Boote, Kenneth J and Thorburn, Peter J and R{\"o}tter, Reimund P and Cammarano, Davide and others},
  journal={Nature climate change},
  volume={3},
  number={9},
  pages={827--832},
  year={2013},
  publisher={Nature Publishing Group UK London}
}

@article{ewert2015uncertainties,
  title={Uncertainties in scaling-up crop models for large-area climate change impact assessments},
  author={Ewert, Frank and van Bussel, Lenny GJ and Zhao, Gang and Hoffmann, Holger and Gaiser, Thomas and Specka, Xenia and Nendel, Claas and Kersebaum, Kurt-Christian and Sosa, Carmen and Lewan, Elisabet and others},
  journal={Handbook of Climate Change and Agroecosystems},
  pages={452--p},
  year={2015},
  publisher={World Scientific Publishing-Imperial College Press}
}

@article{challinor2018improving,
  title={Improving the use of crop models for risk assessment and climate change adaptation},
  author={Challinor, Andrew J and M{\"u}ller, Christoph and Asseng, Senthold and Deva, Chetan and Nicklin, Kathryn Jane and Wallach, Daniel and Vanuytrecht, Eline and Whitfield, Stephen and Ramirez-Villegas, Julian and Koehler, Ann-Kristin},
  journal={Agricultural systems},
  volume={159},
  pages={296--306},
  year={2018},
  publisher={Elsevier}
}

@article{waqar2025stacking,
  title={A stacking ensemble framework leveraging synthetic data for accurate and stable crop yield forecasting},
  author={Waqar, Muhammad and Kim, Yong-Woon and Byun, Yung-Cheol},
  journal={IEEE Access},
  volume={13},
  pages={136909--136926},
  year={2025},
  publisher={IEEE}
}

\end{document}